\documentclass[11pt]{article}

\usepackage[final]{acl}

\usepackage{times}
\usepackage{latexsym}

\usepackage[T1]{fontenc}
\usepackage[utf8]{inputenc}

\usepackage{microtype}

\usepackage{inconsolata}

\usepackage{enumitem}
\usepackage{amsmath}
\usepackage{amsfonts}

\usepackage{hyperref}       % hyperlinks
\usepackage{url}            % simple URL typesetting
\usepackage{booktabs}       % professional-quality tables
\usepackage{nicefrac}       % compact symbols for 1/2, etc.
\usepackage{xcolor}         % colors
\usepackage{amssymb}
\usepackage{mathtools}
\usepackage{multirow}
\usepackage{colortbl}
\usepackage{arydshln}
\usepackage{subfigure}
\usepackage{wrapfig}
\usepackage{algorithm}
\usepackage{algpseudocode}
\usepackage{subcaption}
\usepackage{float}
\usepackage{adjustbox}
\usepackage{graphicx}
\usepackage{amsthm}

\usepackage[capitalize,noabbrev]{cleveref}
\usepackage{bbding}
\usepackage{pifont}
\usepackage{minitoc}
\usepackage{booktabs}
\usepackage{minitoc}

\definecolor{myblue}{RGB}{224,247,250}
\definecolor{myblue2}{RGB}{186,230,251}
\definecolor{mintleaf}{RGB}{0, 184, 148}

\definecolor{backblue}{RGB}{210, 230, 250}

\definecolor{backred}{RGB}{255, 223, 223}

\definecolor{backgreen}{RGB}{220,244,229}

\definecolor{back_deepblue}{RGB}{180, 210, 240}

\definecolor{back_deepred}{RGB}{255, 200, 200}

\definecolor{back_deepgreen}{RGB}{190, 230, 210}

\definecolor{mygray}{gray}{0.95}
\definecolor{greentable3}{rgb}{0,0.5,0}
\definecolor{mgreen}{RGB}{6,128,67}
\definecolor{mgray}{RGB}{128,128,128}
\definecolor{mygreen}{RGB}{233,247,234}

\title{SPARK: Strategic Policy-Aware Exploration via Dynamic Branching for Long-Horizon Agentic Learning}
\author{
 \textbf{Jinyang Wu\textsuperscript{1}\thanks{\quad Equal Contribution.}},
 \textbf{Shuo Yang\textsuperscript{1}\footnotemark[1]},
 \textbf{Changpeng Yang\textsuperscript{2}},
 \textbf{Yuhao Shen\textsuperscript{3}},\\
 \textbf{Shuai Zhang\textsuperscript{1}},
 \textbf{Zhengqi Wen\textsuperscript{1}},
 \textbf{Jianhua Tao\textsuperscript{1}\footnotemark[2]}
\\
 \textsuperscript{1}Tsinghua University\;\;
 \textsuperscript{2}Peking University\;\;
 \textsuperscript{3}Zhejiang University
\\
\texttt{\{wu-jy23, shuo-yan20\}@mails.tsinghua.edu.cn}
}

\begin{document}
\maketitle

% WJY 1212 version
\begin{abstract}
Reinforcement learning has empowered large language models to act as intelligent agents, yet training them for long-horizon tasks remains challenging due to the scarcity of high-quality trajectories, especially under limited resources. Existing methods typically scale up rollout sizes and indiscriminately allocate computational resources among intermediate steps. Such attempts inherently waste substantial computation budget on trivial steps while failing to guarantee sample quality. To address this, we propose \textbf{\textsc{Spark}} (\textbf{S}trategic \textbf{P}olicy-\textbf{A}ware explo\textbf{R}ation via \textbf{K}ey-state dynamic branching), a novel framework that selectively branches at critical decision states for resource-efficient exploration. Our key insight is to activate adaptive branching exploration at critical decision points to probe promising trajectories, thereby achieving precise resource allocation that prioritizes sampling quality over blind coverage. This design leverages the agent's intrinsic decision-making signals to reduce dependence on human priors, enabling the agent to autonomously expand exploration and achieve stronger generalization. Experiments across diverse tasks (e.g., embodied planning), demonstrate that \textsc{Spark} achieves superior success rates with significantly fewer training samples, exhibiting robust generalization even in unseen scenarios. Our code and checkpoints are available at \href{https://github.com/jinyangwu/SPARK}{https://github.com/jinyangwu/SPARK}.
\end{abstract}

\section{Introduction}\label{sec:intro}
Reinforcement Learning (RL) has proven effective in enhancing Large Language Model (LLM) reasoning capabilities, with notable success in domains such as mathematics~\citep{guo2025deepseek,o1} and coding~\citep{zhoubian2025rest}. This has led to a growing consensus that RL holds the key to the next frontier~\citep{zhang2025rlsurvey,yue2025does}: \textit{agentic AI}—systems capable of autonomously navigating dynamic environments and executing complex, long-horizon instructions~\citep{zhang2025landscape,wu2026atlas}.

\begin{figure}[t]
  % \centering
  % \hspace{-0.25cm}
  \includegraphics[width=1.0\linewidth]{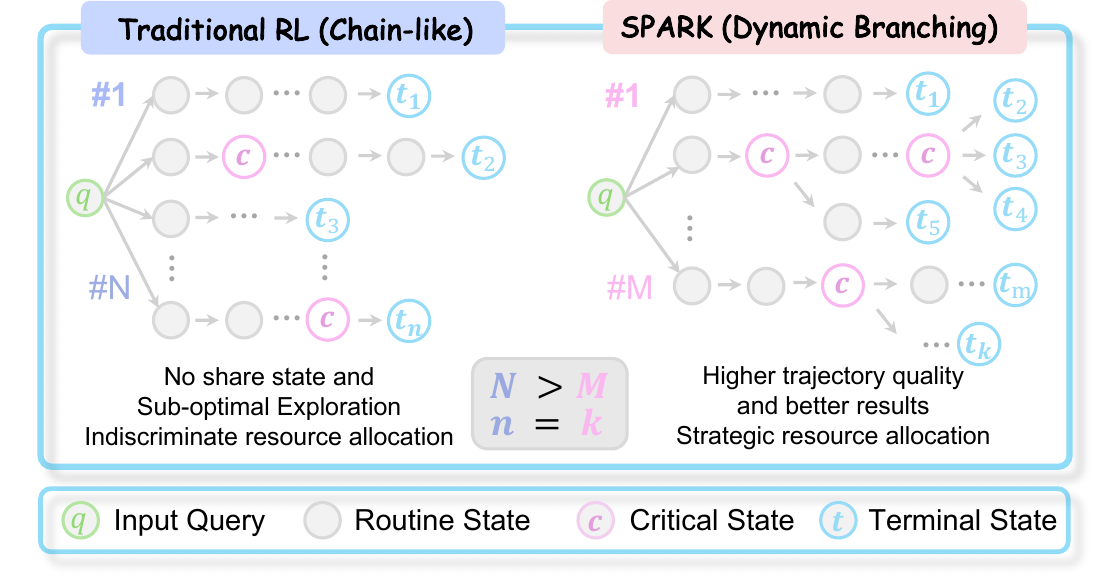}
    \caption{\textbf{Paradigm comparison: uniform vs. strategic exploration.} Standard RL (\textbf{Left}) wastes budget via uniform, independent sampling. In contrast, \textsc{Spark} (\textbf{Right}) employs dynamic branching at critical junctures for precise resource allocation, yielding higher-quality trajectories under the similar computational budget.}
  \label{motivation}
  \vspace{-0.1in}
\end{figure}

However, applying RL to agentic tasks faces a fundamental bottleneck: \textit{the scarcity of high-quality trajectories under resource constraints}. Unlike math problems where solutions are self-contained and directly verifiable~\citep{zhang2025rlsurvey}, agentic tasks require navigating vast state spaces where a single misstep can derail a long sequence~\citep{feng2025groupingroup,zhang2025rlvmr}. Consequently, successful trajectories are sparse, making it difficult for the policy to learn effectively~\citep{kimiresearcher2025,wu2025templaterl}.

Existing methods attempt to mitigate this challenge by expanding exploration budget and employing extensive search-based techniques during RL training~\citep{xing2025lookahead,ji2025tree}. Yet, these approaches suffer from \textit{indiscriminate resource allocation}: they distribute computational budget uniformly across all steps (Figure~\ref{motivation}), rather than targeting decision points where additional exploration is most valuable. Consider an embodied agent tasked with preparing breakfast~\citep{salimpour2025towards,liang2025large}: uniform exploration wastes substantial resources on trivial steps like ``opening the fridge door'' while allocating insufficient budget to pivotal states like ``choosing alternative ingredients when the intended ones are missing.'' By prioritizing exhaustive coverage over sampling quality, such methods fail to yield high-value trajectories under resource constraints, resulting in inefficient exploration and unstable training.

\begin{figure}[t]
  % \centering
  % \hspace{-0.25cm}
  \includegraphics[width=1.0\linewidth]{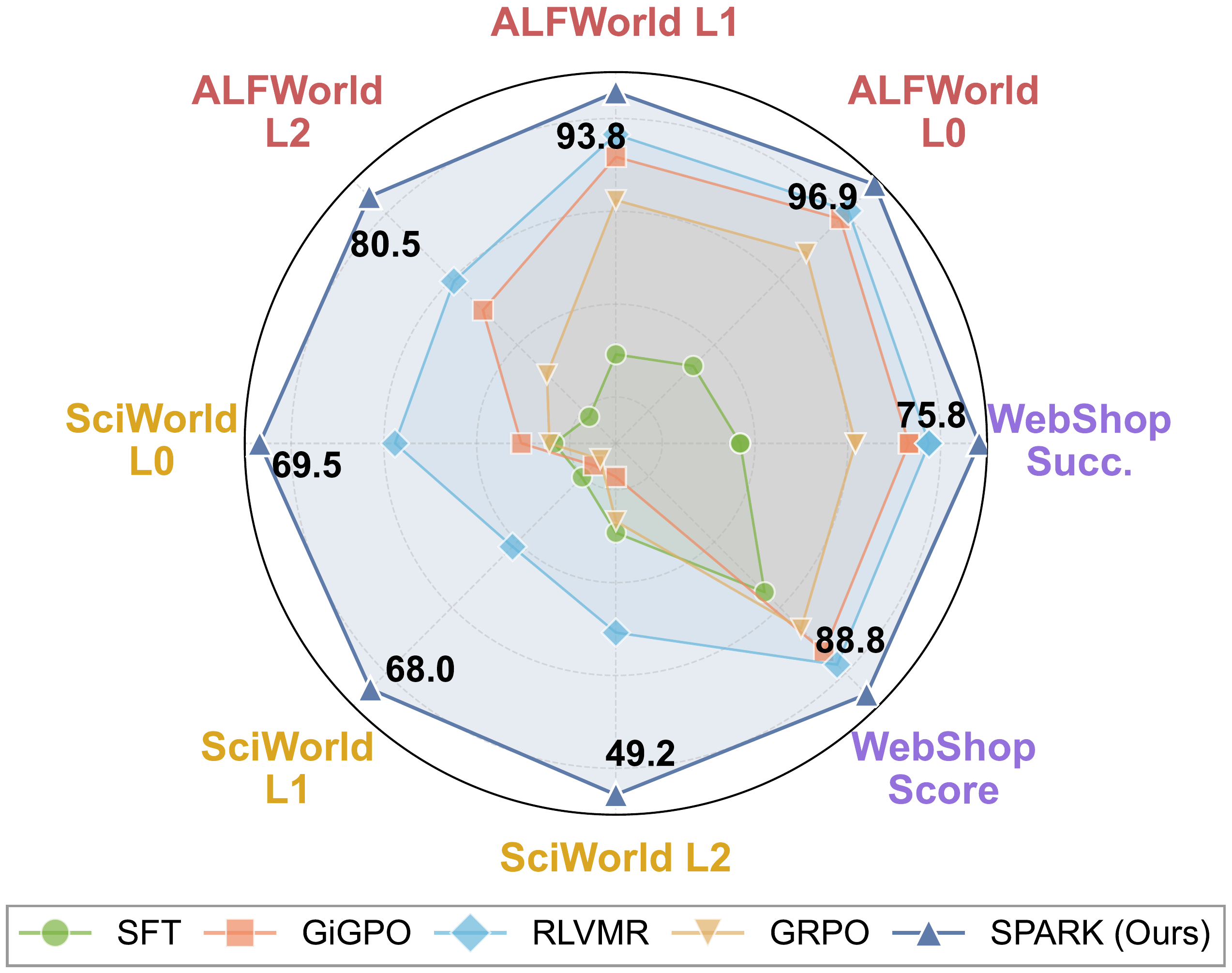}
    \caption{\textbf{Multi-benchmark performance comparison.} SPARK outperforms all baselines across ALFWorld (L0-L2), ScienceWorld (L0-L2), and WebShop tasks, achieving +73.5\% average improvement.}
  \label{overview_performance}
  \vspace{-0.05in}
\end{figure}

We argue that effective exploration should be triggered at appropriate decision points. Rather than blind search, an intelligent agent should autonomously identify critical junctures at which embarking on additional exploration branches yields the greatest potential benefits. Building on this principle, we introduce \textbf{\textsc{Spark}} (\textbf{S}trategic \textbf{P}olicy-\textbf{A}ware explo\textbf{R}ation via \textbf{K}ey-state dynamic branching), a novel framework that enables autonomous strategic exploration via dynamic branching at critical intermediate states, which we refer to as \textsc{Spark} points. Our key insight is to leverage the agent's intrinsic decision-making signals to govern the exploration topology. When encountering states with high epistemic uncertainty or semantic ambiguity, the agent selectively initiates additional exploration; otherwise, it proceeds linearly through routine decisions. This achieves precise resource allocation that prioritizes sampling quality over exhaustive coverage while minimizing reliance on human priors, thereby ensuring robust generalization. Extensive experiments show that \textsc{Spark} significantly enhances performance, efficiency, and out-of-domain generalization over powerful baselines. Our contributions are three-fold:

\begin{itemize}[leftmargin=1.18em]
    \item \textbf{Autonomous Strategic Exploration}: We propose an agentic RL framework that enables agents to autonomously identify intermediate states where additional exploration is needed, relying on intrinsic decision-making signals rather than handcrafted heuristics, thereby enabling precise and efficient resource allocation.
    
    \item \textbf{Adaptive Dynamic Branching}: By dynamically triggering branching at critical states, we achieve superior sample quality under constrained budget, generating more informative training trajectories than blind exploration.
    
    \item \textbf{Compelling Strong Empirical Validation}: Experiments (Figure~\ref{overview_performance}) on challenging tasks (e.g., embodied planning in ALFWorld) demonstrate that \textsc{Spark} achieves superior success rates with higher sample efficiency, exhibiting robust generalization even in unseen scenarios.
\end{itemize}

\begin{figure*}[t]
     \includegraphics[width=\linewidth]{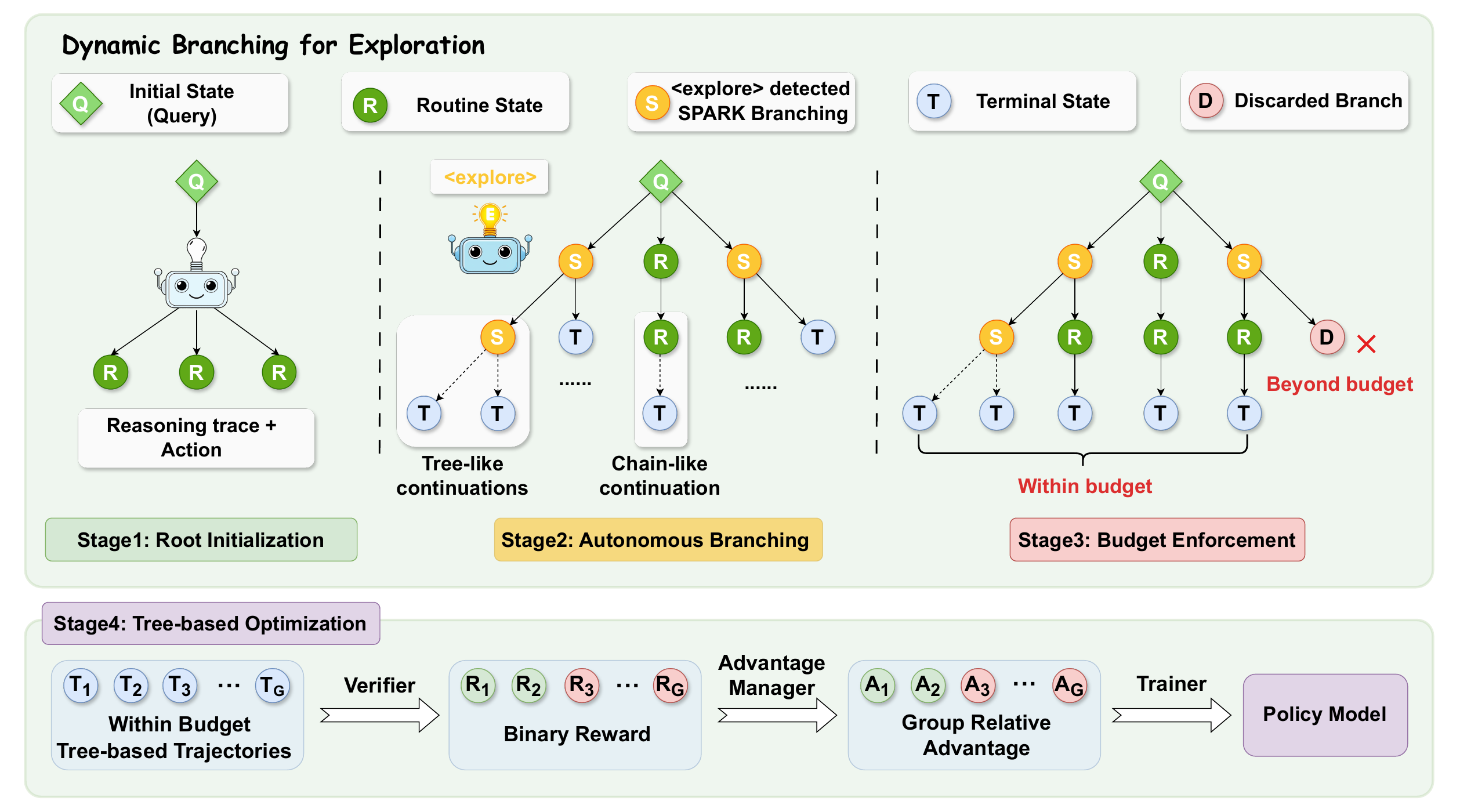}
    \caption{\textbf{Overview of \textsc{Spark} framework.} \textsc{Spark} performs dynamic branching exploration via: (1) \textit{Root Initialization}: diverse starting trajectories; (2) \textit{Autonomous Branching}: selective expansion at high-uncertainty states using intrinsic \texttt{<explore>} signals; (3) \textit{Budget Enforcement}: constraining tree growth within computational limits. The resulting trajectory trees are then used for \textit{Tree-Based Policy Optimization}.}
    \label{fig:beacon}
\end{figure*}

\section{Methodology}\label{sec:method}
We first formulate the agentic task as a partially observable decision process (\S\ref{sec:problem}), then introduce our core mechanism: \textit{Dynamic Branching Exploration} that selectively allocates exploration budget based on intrinsic decision-making signals (\S\ref{sec:branching}). Figure~\ref{fig:beacon} provides the overview of \textsc{Spark}.
% In this section, we present \textsc{Spark}, a framework that enables autonomous strategic exploration through dynamic branching. We first formulate the agentic task as a partially observable decision process (\S\ref{sec:problem}). We then introduce our core mechanism: \textit{Dynamic Branching} that selectively allocates exploration resources based on the agent's intrinsic decision-making signals (\S\ref{sec:branching}), followed by the \textit{Tree-Based Policy Optimization} (\S\ref{sec:training}). Figure~\ref{fig:beacon} presents an overview of the \textsc{Spark} pipeline. 

\subsection{Problem Formulation}\label{sec:problem}
We formulate long-horizon agentic tasks as a Partially Observable Markov Decision Process (POMDP), defined by the tuple $(\mathcal{S}, \mathcal{A}, \mathcal{O}, \mathcal{T}, \mathcal{R}, \gamma)$, where $\mathcal{S}$ is the latent state space, $\mathcal{A}$ is the action space, $\mathcal{O}$ is the observation space, $\mathcal{T}: \mathcal{S} \times \mathcal{A} \rightarrow \mathcal{S}$ is the transition function, $\mathcal{R}: \mathcal{S} \times \mathcal{A} \rightarrow \mathbb{R}$ is the reward function, and $\gamma \in [0,1)$ is discount factor.

At each timestep $t$, the environment is in a hidden state $s_t \in \mathcal{S}$. The agent receives observation $o_t \in \mathcal{O}$ (e.g., textual feedback), and maintains an interaction history $h_t = (o_0, z_0, a_0, \dots, o_t)$, where $z_i$ denotes the reasoning trace and $a_i \in \mathcal{A}$ is the action at step $i$. This history serves as the belief state proxy. Then, the agent (policy model $\pi_\theta$) generates a reasoning trace $z_t$ followed by an action $a_t$:
\begin{equation}\label{equ1}
    (z_t, a_t) \sim \pi_\theta(\cdot \mid h_t).
\end{equation}

Upon executing action $a_t$, the environment transitions to $s_{t+1}$ according to $\mathcal{T}(s_{t+1} \mid s_t, a_t)$ and provides observation $o_{t+1}$. We represent a single step as the triplet $\tau_t = (o_t, z_t, a_t)$. An episode continues until the agent achieves the goal or reaches the maximum horizon $K$, yielding a trajectory $\tau = (\tau_0, \tau_1, \dots, \tau_T)$ where $T \leq K$. 

For most agentic benchmarks~\citep{singh2025agentic}, the reward function provides sparse terminal feedback: $R(\tau) \in \{0, 1\}$ indicating task success or failure. Our objective is to find an optimal policy $\pi^*$ that maximizes the expected return:
\begin{equation}\label{equ2}
    J(\pi) = \mathbb{E}_{\tau \sim \pi} [R(\tau)].
\end{equation}

\vspace{-0.1in}
% \subsection{Resource-Aware Dynamic Branching \textcolor{red}{Dynamic Branching Exploration} }\label{sec:branching}
\subsection{Dynamic Branching Exploration}\label{sec:branching}
Uniform exploration across all steps is highly inefficient in long-horizon RL~\citep{wang2020longRl}. Trivial steps (e.g., open the door in front of a fridge) require minimal search, while critical junctures (e.g., choosing alternative ingredients when the intended ones are missing) benefit substantially from exploring multiple alternatives. Standard methods allocate resources uniformly, wasting computation on routine actions while under-exploring pivotal decisions. We address this through \textit{dynamic branching}: a mechanism that constructs \textbf{hierarchical tree-structured exploration paths} by selectively allocating rollout budget at \textbf{pivotal decision points (\textsc{Spark} points)}, guided by the agent's intrinsic decision-making signals.
%, manifested as an explicit <explore> tag in reasoning trace. 
Supervised fine-tuning (SFT) is first employed to endow the model with the preliminary capability to generate such intrinsic signals, followed by RL optimization.

Let $N$ denote the total rollout budget (maximum leaf nodes) and $K$ denote the maximum episode length. Rather than sampling $N$ independent trajectories, \textsc{Spark} builds a trajectory forest that branches at states where additional exploration is valuable. Specifically, we operate in three stages:

\begin{algorithm*}[t!]
\caption{\textsc{Spark}: Strategic Policy-Aware Exploration via Key-State Dynamic
Branching}
\renewcommand{\algorithmicrequire}{\textbf{Input:}}
\renewcommand{\algorithmicensure}{\textbf{Output:}}
\label{alg:beacon}
\begin{algorithmic}[1]
    \Require Policy $\pi_{\theta}$; task dataset $\mathcal{Q}$; rollout budget $N$; branching factor $B$; initial roots $M$
    \Ensure Optimized policy $\pi_{\theta}^*$
    
    \For{iteration $= 1, 2, \dots$}
        % \State \textcolor{mintleaf}{// Resource-Aware Dynamic Branching}
        \State $\mathcal{D} \gets \emptyset$ 
        \For{task $q \sim \mathcal{Q}$}
            \Comment{\textcolor{mintleaf}{\textit{Dynamic Branching Exploration}}}
            % \tcp*[r]{\textcolor{mintleaf}{\textit{Exploration}}}
            \State Initialize $M$ parallel trajectory roots from $q$
            \For{each active trajectory $i$ at step $t$}
                \State Generate reasoning trace and action $(z_t^{(i)},a_t^{(i)}) \sim \pi_\theta(\cdot \mid h_t^{(i)})$
                \State Compute branching factor $b_t^{(i)} \gets \mathcal{B}(z_t^{(i)})$ upon triggering \texttt{<explore>} in $z_t^{(i)}$
                \State Get effective branching factor $b_{t,\text{eff}}^{(i)}$ under budget constraint $N$ via Eq. \eqref{b_eff}
                \State Prepare context for $b_{t,\text{eff}}^{(i)}$ continuations and update active trajectory mask
            \EndFor

            \State Collect completed trajectory group $G$ into $\mathcal{D}$
        \EndFor
        \State Update $\theta$ using GRPO objective with collected trajectories \Comment{\textcolor{mintleaf}{\textit{Tree-Based Policy Update}}}
    \EndFor
\end{algorithmic}
\end{algorithm*}

\vspace{-0.05in}
\paragraph{Stage 1: Root Initialization.}
At initial state $s_0$, task ambiguity is typically highest. To ensure exploration diversity, we create $M$ parallel trajectory roots by sampling distinct reasoning-action pairs:
\begin{equation}
    \{(z_0^{(i)}, a_0^{(i)})\}_{i=1}^M \sim \pi_\theta(\cdot \mid h_0),
\end{equation}
where $M < N$ (typically $M = 3$ to $5$). This establishes $M$ independent trajectory trees.

\paragraph{Stage 2: Autonomous Branching.}
For each active trajectory $i$ at step $t$, the agent generates a reasoning trace $z_t^{(i)}$ that internally deliberates whether the current decision requires additional exploration. We employ a branching criterion $\mathcal{B}(\cdot)$ that analyzes $z_t^{(i)}$ to determine exploration intensity:
\begin{equation}
    b_t^{(i)} = \mathcal{B}(z_t^{(i)}),
\end{equation}
where $b_t^{(i)} \in \{1, B\}$ specifies the branching factor—either linear continuation ($b=1$) or multi-branch exploration ($b=B$).

\underline{\textit{Branching Criterion.}} The function $\mathcal{B}$ identifies states where exploration is valuable by detecting signals of epistemic uncertainty in the reasoning trace, which is explicitly manifested as a dedicated \texttt{<explore>} tag.
% These signals include: (1) explicit uncertainty markers (e.g., ``I'm unsure which...'', ``need to explore...''), (2) enumeration of multiple options (e.g., ``could try A or B''), or (3) error recovery patterns (e.g., ``the previous action failed...''). In practice, we implement $\mathcal{B}$ using simple pattern matching on the reasoning text, though learned classifiers could be employed. 
When uncertainty is detected, we set $b_t^{(i)} = B$; otherwise $b_t^{(i)} = 1$.  Function $\mathcal{B}$ can be formalized as:
\begin{equation}
    \label{b_function}
    \mathcal{B}(z_t^{(i)})=
    \begin{cases} 
    B, \quad\text{if \texttt{<explore>} in } z_t^{(i)} \\
    1, \quad\text{ otherwise.}
    \end{cases}
\end{equation}
Based on $b_t^{(i)}$, we sample continuations:
\begin{equation}
    \{(z_{t+1}^{(i,j)}, a_{t+1}^{(i,j)})\}_{j=1}^{b_t^{(i)}} \sim \pi_\theta(\cdot \mid h_{t+1}^{(i)}),
\end{equation}
where $h_{t+1}^{(i)} = (h_t^{(i)}, z_t^{(i)}, a_t^{(i)}, o_{t+1}^{(i)})$ is the updated history. This creates $b_t^{(i)}$ child trajectories from the current node.

\paragraph{Stage 3: Budget Enforcement.} Expansion is constrained by the global budget $N$. Let $N_{\text{current}}$ denote the current number of active leaf nodes. The actual branching factor is adjusted dynamically:
\begin{equation}
    \label{b_eff}
    b_{t,\text{eff}}^{(i)} 
    =\min(b_t^{(i)}, N - N_{\text{current}} + 1).
\end{equation}
This ensures we never exceed budget $N$. Rollout terminates when a trajectory reaches a terminal state (success/failure) or maximum horizon $K$.

This process yields a collection of $M$ trajectory trees containing $|G|$ completed trajectories, where $M \leq |G| \leq N$. We denote this collection as a \textit{group} $G = \{\tau^{(1)}, \tau^{(2)}, \dots, \tau^{(|G|)}\}$.

\paragraph{Stage 4: Tree-Based Policy Update.}
The trajectory trees generated by \textsc{Spark} are used to update the policy without modifying the underlying optimization algorithm.
Completed leaf trajectories from the same task are grouped and assigned binary terminal rewards, which are propagated to all steps.
Shared prefixes induce comparable alternatives, enabling relative credit assignment via group-normalized advantages.
Policy updates are performed using standard clipped optimization with KL term against a reference policy, remaining compatible with existing GRPO-style pipelines while incorporating decision-focused exploration signals.

\subsection{Theoretical Perspective on \textsc{Spark}}
\label{sec:theory}

We present a concise theoretical argument explaining why \textsc{Spark} achieves
substantially improved performance under \emph{consistently lower token budgets}
in long-horizon decision-making tasks.

\paragraph{Key Insight.}
Let $K$ denote the episode horizon and $\mathcal{C} \subseteq \{0,\dots,K-1\}$ the set
of \emph{pivotal decision steps} (\textsc{Spark} points) whose action choices critically affect
task success, with $|\mathcal{C}| = m \ll K$.
Standard GRPO-style linear sampling allocates computation uniformly across all steps
of a trajectory.
In contrast, \textsc{Spark} redistributes intermediate computation toward pivotal
steps without increasing the final number of completed trajectories.

\paragraph{Decision Coverage at Pivotal Steps.}
At a pivotal step $t \in \mathcal{C}$, let $q_t \in (0,1)$ denote the probability mass
assigned by the current policy to desirable actions when sampled once.
Under linear sampling, each trajectory attempts this decision only once, yielding
a coverage probability of $q_t$.

In \textsc{Spark}, dynamic branching generates $B$ conditional continuations at
pivotal steps, increasing the probability of selecting at least one desirable action
to
\begin{equation}
    q_t^{\text{branch}} = 1 - (1 - q_t)^B,
\end{equation}
which strictly dominates $q_t$ for any $B \ge 2$.

\paragraph{Implications for Long-Horizon Tasks.}
Task success typically depends on a sequence of correct decisions at pivotal steps.
By replacing each $q_t$ with $q_t^{\text{branch}}$, \textsc{Spark} amplifies the
probability of discovering high-quality trajectories through a multiplicative effect
across $\mathcal{C}$.
Meanwhile, \textsc{Spark} avoids repeatedly rolling out suffixes induced by
suboptimal pivotal decisions, leading to fewer redundant computations at
non-pivotal steps and a modest reduction in overall token usage.

\section{Experiments}
\label{sec:experiments}
We analyze our framework from four aspects: (1) \textbf{Performance}: superior success rates versus strong baselines; (2) \textbf{Efficiency}: sample and token efficiency; (3) \textbf{Generalization}: robustness on unseen scenarios; (4) \textbf{Ablation Study}: contribution of dynamic branching and exploration quality analysis.

\subsection{Experimental Settings}
\label{subsec:settings}
\paragraph{Benchmarks.}
Following prior work~\citep{feng2025groupingroup,zhang2025rlvmr}, we evaluate on three challenging long-horizon agentic domains: (1) \textbf{Embodied Decision Making}: ALFWorld~\citep{shridhar2020alfworld} for household task execution; (2) \textbf{Scientific Reasoning}: ScienceWorld~\citep{scienceworld} with complex task horizons up to 30+ steps; (3) \textbf{Web Navigation}: WebShop~\citep{yao2022webshop} with 1.1M products to test open-ended exploration.

\paragraph{Baselines.}
We compare \textsc{Spark} against: (1) \textbf{Closed-source LLMs}: GPT-4o~\citep{gpt4o}, GPT-5-mini~\citep{gpt5}, GPT-5~\citep{gpt5}, Gemini-2.5-Pro~\citep{comanici2025gemini}; (2) \textbf{Prompting Methods}: ReAct~\citep{yao2022react}; and (3) \textbf{RL Methods}: GRPO~\citep{shao2024deepseekmath}, ETO~\citep{song2024trial}, GiGPO~\citep{feng2025groupingroup}, and RLVMR~\citep{zhang2025rlvmr}.

\paragraph{Evaluation Metrics.}
Following~\citet{shridhar2020alfworld,yao2022webshop}, we report \textit{Success Rate (SR)} for task completion and \textit{Average Score} (WebShop only) for partial success (e.g., selecting a product with correct attributes but wrong color).

% \vspace{-0.05in}
\paragraph{Implementation Details.}
We use Qwen2.5-1.5B/7B-instruct~\citep{qwen25} as base models. A cold-start SFT with 300 trajectories enables initial \texttt{<explore>} tag generation in RL. We set $N=8$, $M=4$, $B=2$ for rollout budget, initial roots, and branching factor, respectively. RL training runs for 120 steps with batch size 16. Full details are provided in Appendix~\ref{subsec:implementation}.

\begin{table*}[t]
\centering
\vspace{-0.05in}
\resizebox{1.0\linewidth}{!}{
\begin{tabular}{lccccccccccc}
\toprule
\multirow{2}{*}{\textbf{Method}} & \multicolumn{6}{c}{\textbf{ALFWorld}} & \multicolumn{3}{c}{\textbf{ScienceWorld}} & \multicolumn{2}{c}{\textbf{WebShop}} \\
\cmidrule(lr){2-7} \cmidrule(lr){8-10} \cmidrule(lr){11-12}
& Look & Clean & Pick2 & L0 & L1 & L2 & L0 & L1 & L2 & score & succ. \\
\midrule	
\rowcolor{gray!8}\multicolumn{12}{c}{\textit{Closed-Source Models}}\\
GPT-4o & 50.0 & 34.4 & 15.8 & 52.3 & 48.4 & 38.3 & 32.0 & 32.8 & 32.8 & 12.7 & 6.2 \\
GPT-5-mini & 58.3 & 28.1 & 52.6 & 57.0 & 54.7 & 46.1 & 39.8 & 31.3 & 31.4 & 14.2 & 7.0 \\
GPT-5 & 75.0 & 43.8 & 57.9 & 70.0 & 60.2 & 63.3 & 53.9 & 33.6 & 33.6 & 34.9 & 29.7 \\
Gemini-2.5-Pro & 75.0 & 40.6 & 84.2 & 56.3 & 55.0 & 55.5 & 32.8 & 27.3 & 30.5 & 42.9 & 32.0 \\
\midrule
\midrule
\rowcolor{gray!8}\multicolumn{12}{c}{\textit{Advanced Method Comparison}}\\
\multicolumn{12}{l}{\textit{Qwen2.5-1.5B}} \\
\;\; ReAct & 18.3 & 10.8 & 0.0 & 11.3 & 13.7 & 10.2 & 1.2 & 0.8 & 0.8 & 40.1 & 11.3 \\
\;\; SFT & 42.0 & 35.9 & 29.7 & 43.0 & 38.7 & 17.6 & 20.3 & 18.0 & 12.5 & 65.8 & 39.1 \\
\;\; ETO & 46.3 & 66.2 & 55.6 & 64.1 & 66.4 & 25.8 & 39.1 & 22.7 & 15.6 & - & - \\
\;\; GiGPO & 66.5 & 90.7 & 73.8 & 86.7 & 83.2 & 48.0 & 25.8 & 15.2 & 4.7 & 83.1 & 65.0 \\
\;\; RLVMR & 78.8 & 91.2 & 77.6 & 89.1 & 87.9 & 56.3 & 46.9 & 34.4 & 26.5 & 86.7 & 71.1 \\
\;\; GRPO & 55.6 & 88.1 & 72.1 & 76.6 & 71.1 & 29.7 & 21.1 & 13.7 & 10.9 & 75.8 & 56.8 \\
\rowcolor[RGB]{236,244,252}
\;\; \textsc{Spark} (Ours) & \textbf{91.7} & \textbf{100.0} & \textbf{89.5} & \textbf{96.9} & \textbf{93.8} & \textbf{80.5} & \textbf{69.5} & \textbf{68.0} & \textbf{49.2} & \textbf{88.8} & \textbf{75.8} \\
\rowcolor{pink!15}
\;\; $\bigtriangleup$ vs. GRPO & +36.1 & +11.9 & +17.4 & +20.3 & +22.7 & +50.8 & +48.4 & +54.3 & +38.3 & +13.0 & +19.0 \\ 
\midrule
\multicolumn{12}{l}{\textit{Qwen2.5-7B}} \\
\;\; ReAct & 33.2 & 18.7 & 12.8 & 23.1 & 28.5 & 27.0 & 7.8 & 11.3 & 6.3 & 46.2 & 19.5 \\
\;\; SFT & 63.0 & 61.1 & 33.2 & 63.3 & 57.0 & 37.5 & 36.7 & 32.0 & 23.4 & 62.2 & 38.3 \\
\;\; ETO & 70.5 & 82.3 & 51.2 & 70.3 & 74.2 & 51.6 & 62.5 & 40.6 & 28.1 & - & - \\
\;\; GiGPO & 85.9 & 93.3 & 83.6 & 89.5 & 90.2 & 67.2 & 53.4 & 35.2 & 25.8 & 84.4 & 72.8 \\
\;\; RLVMR & 88.2 & 90.1 & 86.7 & 91.4 & 91.8 & 83.6 & 67.2 & 43.0 & 32.2 & 87.3 & 74.2 \\
\;\; GRPO & 76.7 & 86.0 & 56.4 & 79.3 & 77.3 & 52.3 & 49.1 & 30.1 & 26.6 & 79.3 & 66.1 \\
\rowcolor[RGB]{236,244,252}
\;\; \textsc{Spark} (Ours) & \textbf{100.0} & \textbf{92.5} & \textbf{95.8} & \textbf{96.1} & \textbf{92.2} & \textbf{88.3} & \textbf{75.0} & \textbf{74.1} & \textbf{57.8} & \textbf{89.8} & \textbf{82.8} \\
\rowcolor{pink!15}
\;\; $\bigtriangleup$ vs. GRPO & +23.3 & +6.5 & +39.4 & +16.8 & +14.9 & +36.0 & +25.9 & +44.0 & +31.2 & +10.5 & +16.7 \\ 
\bottomrule
\end{tabular}
}
\caption{\textbf{Performance comparison on long-horizon agentic tasks.} We evaluate success rates (\%) on ALFWorld and ScienceWorld, and task completion score/success rate on WebShop. \textbf{L0/L1/L2} denote seen task categories and instance variants, seen task categories but unseen instance variants, unseen task categories and instance variants, respectively. Look, Clean, and Pick2 represent distinct task types in ALFWorld. Best results per model are in \textbf{bold}.}
\label{table:main_results}
\end{table*}

\subsection{Main Results}\label{subsec:main_results}
Table~\ref{table:main_results} presents results across long-horizon agentic planning tasks, revealing three key findings:

\ding{71} \textit{\textsc{Spark} establishes leading performance across model scales and task domains.} \textsc{Spark} consistently outperforms prompting (ReAct), SFT, and RL methods (GiGPO, RLVMR) on both 1.5B and 7B backbones. Notably, \textsc{Spark}-1.5B attains 49.2\% on the hardest ScienceWorld L2, exceeding GPT-5 (33.6\%) and Gemini-2.5-Pro (30.5\%). This reveals that strategic exploration enables smaller models to rival significantly larger proprietary systems.

\ding{71} \textit{Dynamic branching at critical states substantially outperforms uniform exploration.} Compared to GRPO's uniform sampling, \textsc{Spark} yields substantial gains on 7B: +23.3\% on ALFWorld Look and +39.4\% on Pick2. The amplified improvement on Pick2 (multi-step coordination) versus Look (simple navigation) validates our hypothesis: selective branching allocates budget to critical steps where additional search provides maximum value.

\ding{71} \textit{\textsc{Spark} exhibits robust generalization to unseen environments.} On L2 out-of-domain tasks, \textsc{Spark}-1.5B significantly outperforms GRPO on ALFWorld (80.5\% vs. 29.7\%). On the highly complex ScienceWorld L2, \textsc{Spark} (49.2\%) establishes a 10.5$\times$ lead over GiGPO (4.7\%) and 1.9$\times$ over RLVMR (26.5\%). This highlights the effectiveness of autonomous strategic exploration in novel scenarios where conventional methods struggle.

\subsection{Exploration Efficiency Analysis}\label{subsec:efficiency}
\begin{figure}[t]
  \vspace{-0.05in}
  \includegraphics[width=\columnwidth]{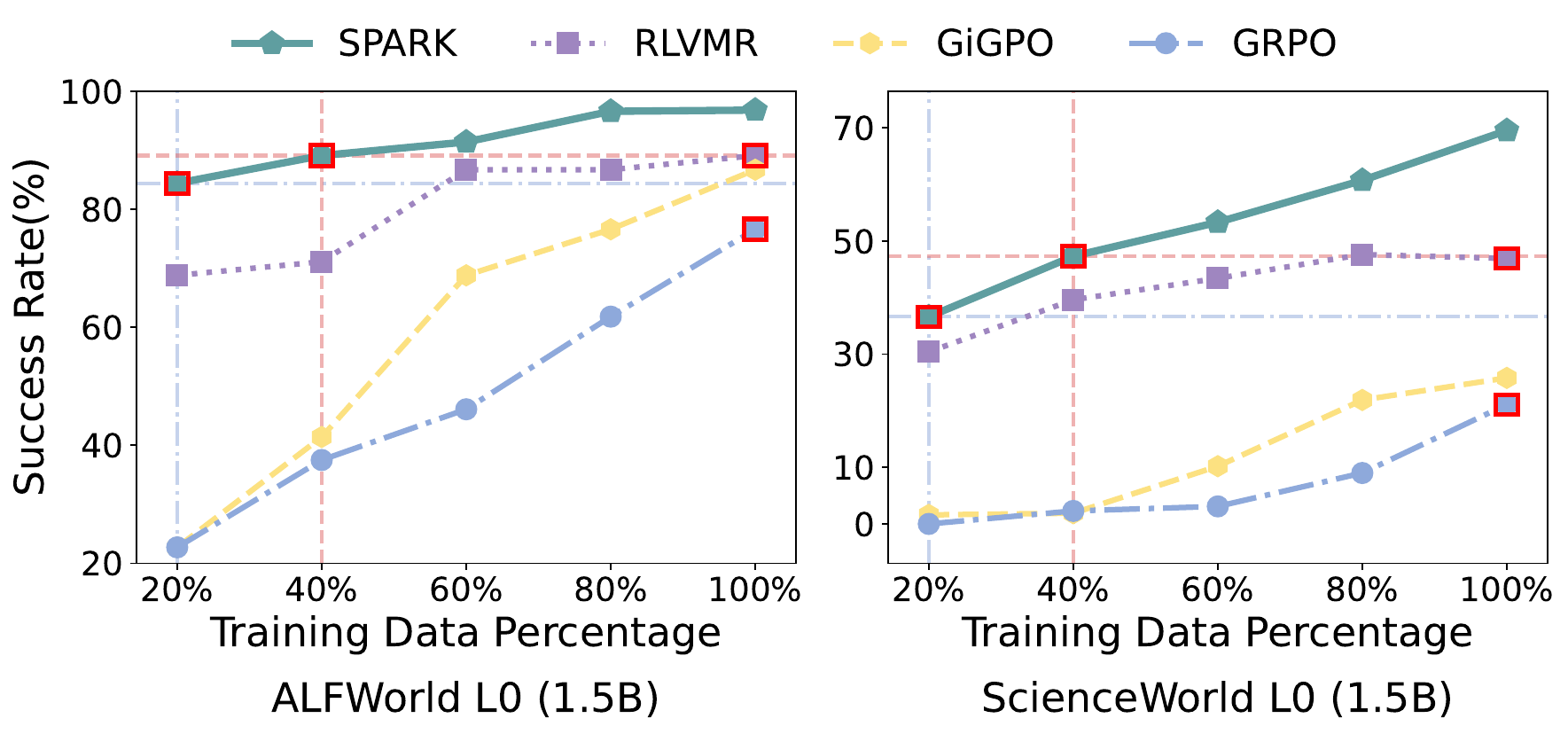}
  \caption{\textbf{Sample efficiency comparison.} \textsc{Spark} surpasses GRPO@100\% using only 20\% data, while baselines often collapse in low-data regimes.}
  \label{fig:data_efficiency}
  % \vspace{-0.2in}
\end{figure}

\paragraph{Sample Efficiency.}
To assess whether \textsc{Spark}'s strategic exploration enhances sample efficiency, we evaluate performance under varying data scales on ALFWorld and ScienceWorld L0 (1.5B). As shown in Figure~\ref{fig:data_efficiency}, \textsc{Spark} demonstrates dramatic efficiency advantages. For example, on ALFWorld, with only 20\% training data, \textsc{Spark} achieves 84.4\% success rate, surpassing GRPO's 76.6\% at full data (100\%). At 40\% data, \textsc{Spark} reaches 89.1\%, matching RLVMR and exceeding GiGPO (86.7\%) at full data. In contrast, baselines collapse at low data regimes: GRPO and GiGPO achieve only 22.7\% at 20\% data, revealing their inability to efficiently learn from limited samples. This 5× data reduction over GRPO confirms that prioritizing exploration at critical decision points enables higher sample efficiency (details in Appendix~\ref{subsec:sample_efficiency}).

\vspace{-0.1in}
\paragraph{Token Efficiency.}
Table~\ref{tab:token_efficiency} compares token consumption of \textsc{Spark} against chain-like methods. By sharing common prefixes across trajectories, our method substantially reduces token generation: 6.9\%, 47.0\%, and 11.2\% on ALFWorld, ScienceWorld, and WebShop, respectively. These results validate our core thesis: by concentrating the exploration budget at critical decision points rather than uniformly across all steps, dynamic branching not only improves sample quality but also eliminates redundant token generation on routine actions, thereby maximizing both performance and computational efficiency with limited resources.

\begin{table}[t!]
    \centering
    \resizebox{1.0\linewidth}{!}{
        \begin{tabular}{lccc}
        \toprule
        Method & ALFWorld$\downarrow$ & ScienceWorld$\downarrow$ & WebShop$\downarrow$ \\
        \midrule
        GRPO & 100.0 & 100.0 & 100.0 \\
        \rowcolor[RGB]{236,244,252}
        \textsc{Spark} & \textbf{93.1} & \textbf{53.0} & \textbf{88.8} \\
        \rowcolor{pink!15}
        Reduction & 6.9 & 47.0 & 11.2 \\
        \bottomrule
        \end{tabular}
    }
    \caption{\textbf{Token efficiency comparison on 1.5B backbone.} We report the relative token consumption of \textsc{Spark} compared to chain-like methods, with the latter normalized to 100 for clear comparison.}
    \label{tab:token_efficiency}
    \vspace{-0.15in}
\end{table}

\begin{table}[ht]
  \centering
  \resizebox{1.0\linewidth}{!}{
  \begin{tabular}{lcccccccc}
    \toprule
    \multirow{2}{*}{\textbf{Method}} & \multicolumn{3}{c}{\textbf{ALFWorld}} & & \multicolumn{3}{c}{\textbf{ScienceWorld}} \\
    \cmidrule(lr){2-4} \cmidrule(lr){6-8}
    & ID $\uparrow$ & OOD $\uparrow$ & $\Delta \downarrow$ & & ID $\uparrow$ & OOD $\uparrow$ & $\Delta \downarrow$ \\
    \midrule
    SFT & 43.0 & 17.6 & \cellcolor[RGB]{236,244,252}59.1\% & & 20.3 & 12.5 & \cellcolor[RGB]{236,244,252}38.4\% \\
    GRPO & 76.6 & 29.7 & \cellcolor[RGB]{236,244,252}61.2\% & & 21.1 & 10.9 & \cellcolor[RGB]{236,244,252}48.3\% \\
    \rowcolor{pink!20}
    \textbf{Ours} & \textbf{96.9} & \textbf{80.5} & \cellcolor[RGB]{236,244,252}\textbf{16.9\%} & & \textbf{69.5} & \textbf{49.2} & \cellcolor[RGB]{236,244,252}\textbf{29.2\%} \\
    \bottomrule
  \end{tabular}}
    \caption{\textbf{Cross-domain generalization comparison.} $\Delta$ denotes the performance drop ratio from seen (in-domain, ID) to unseen scenarios (out-of-domain, OOD).}
  \label{tab:results}
  \vspace{-0.1in}
\end{table}

\vspace{-0.05in}
\subsection{Cross-Domain Generalization}\label{subsec:generalization}
To further evaluate generalization to unseen scenarios, we analyze performance degradation from in-domain (ID, L0) to out-of-domain (OOD, L2) splits. As shown in Table~\ref{tab:results}, GRPO exhibits severe overfitting with performance drops of 61.2\% on ALFWorld and 48.3\% on ScienceWorld. In contrast, \textsc{Spark} maintains robust performance, limiting degradation to 16.9\% and 29.2\% respectively. Moreover, \textsc{Spark} achieves the highest absolute OOD success rates (80.5\% and 49.2\%), substantially outperforming GRPO (29.7\% and 10.9\%). These results indicate that \textsc{Spark}'s dynamic branching enables effective learning of transferable exploration strategies rather than memorization of task-specific trajectories, yielding policies that generalize well to novel environments.

\subsection{Ablation Study and Analysis}\label{subsec:ablation}
\paragraph{Impact of Dynamic Branching.}
To verify dynamic branching, we compare \textsc{Spark} against a ``Fixed Probability'' baseline that randomly triggers branching at every step with fixed probability, independent of decision criticality. As shown in Table~\ref{tab:dynamic_branching}, stochastic branching degrades performance across all datasets, most severely on ScienceWorld (69.5\%→45.3\%) where long horizons amplify the costs of misallocated exploration. This validates our thesis: effective exploration requires \textit{selective} branching at high-uncertainty states rather than uniform expansion, as fixed methods waste budget on trivial steps while under-exploring critical states.

\begin{table}[t]
    \centering
    \resizebox{1.0\linewidth}{!}{
        \begin{tabular}{lccc}
        \toprule
        Method & ALFWorld$\uparrow$ & ScienceWorld$\uparrow$ & WebShop$\uparrow$ \\
        \midrule
        \rowcolor{gray!8}
        Dynamic & \textbf{96.9} & \textbf{69.5} & \textbf{75.8} \\
        Fixed & 90.6 & 45.3 & 70.3 \\
        \rowcolor[RGB]{236,244,252}
        $\bigtriangleup$ & -6.3 & -24.2 & -5.5 \\
        \bottomrule
        \end{tabular}
    }
    \caption{\textbf{Ablation on RL branching strategy.} We report success rates (\%) on 1.5B model. Fixed branching randomly triggers expansion at each step with constant probability, independent of state uncertainty.}
    \label{tab:dynamic_branching}
    \vspace{-0.05in}
\end{table}

\begin{figure}[t]
  \includegraphics[width=\columnwidth]{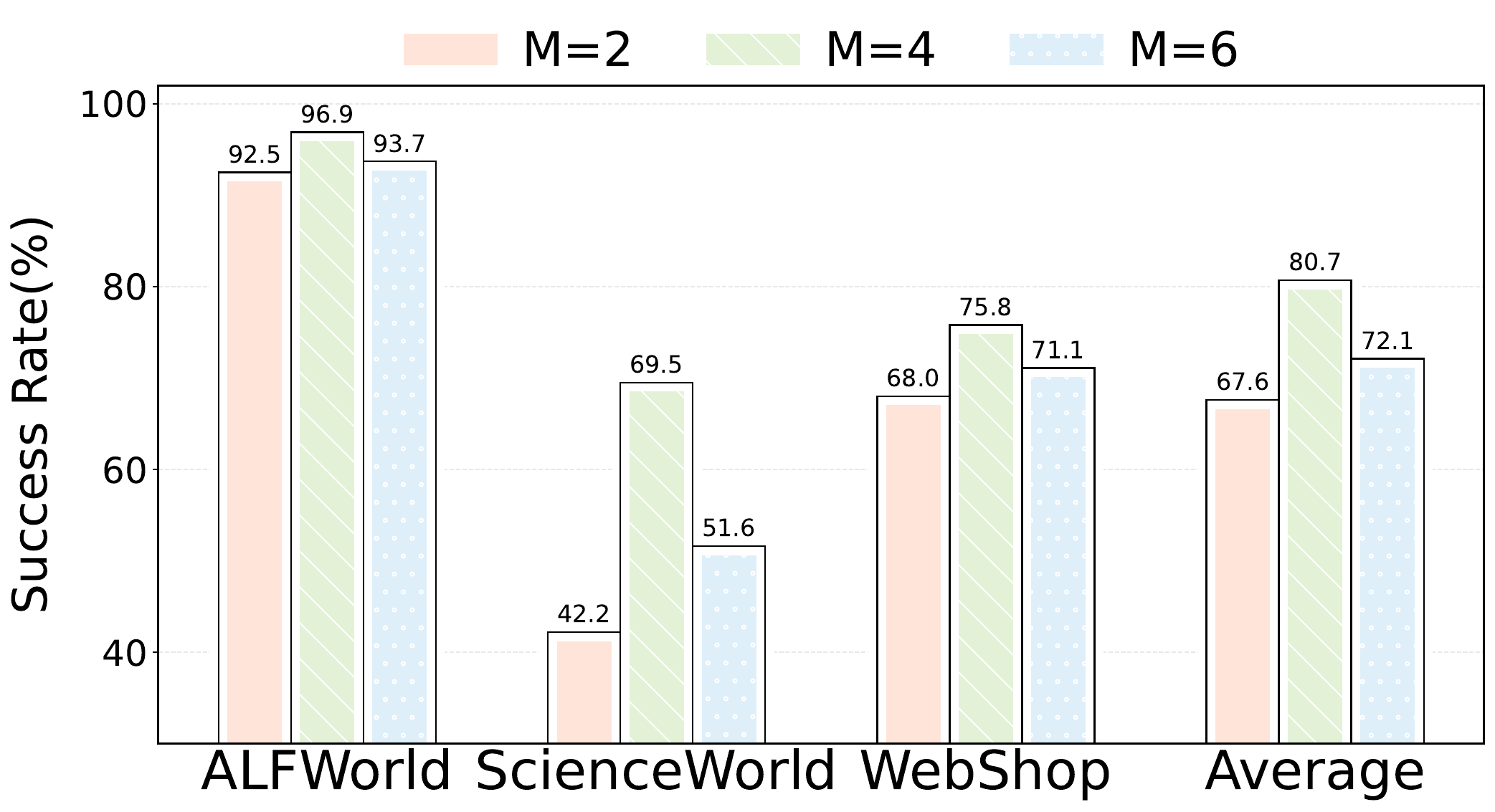}
  \caption{\textbf{Ablation on initial root count ($M$).} We report success rates (\%) with fixed total budget $N=8$.}
  \label{fig:initial_roots}
  \vspace{-0.1in}
\end{figure}

\begin{figure*}[t]
    \includegraphics[width=\linewidth]{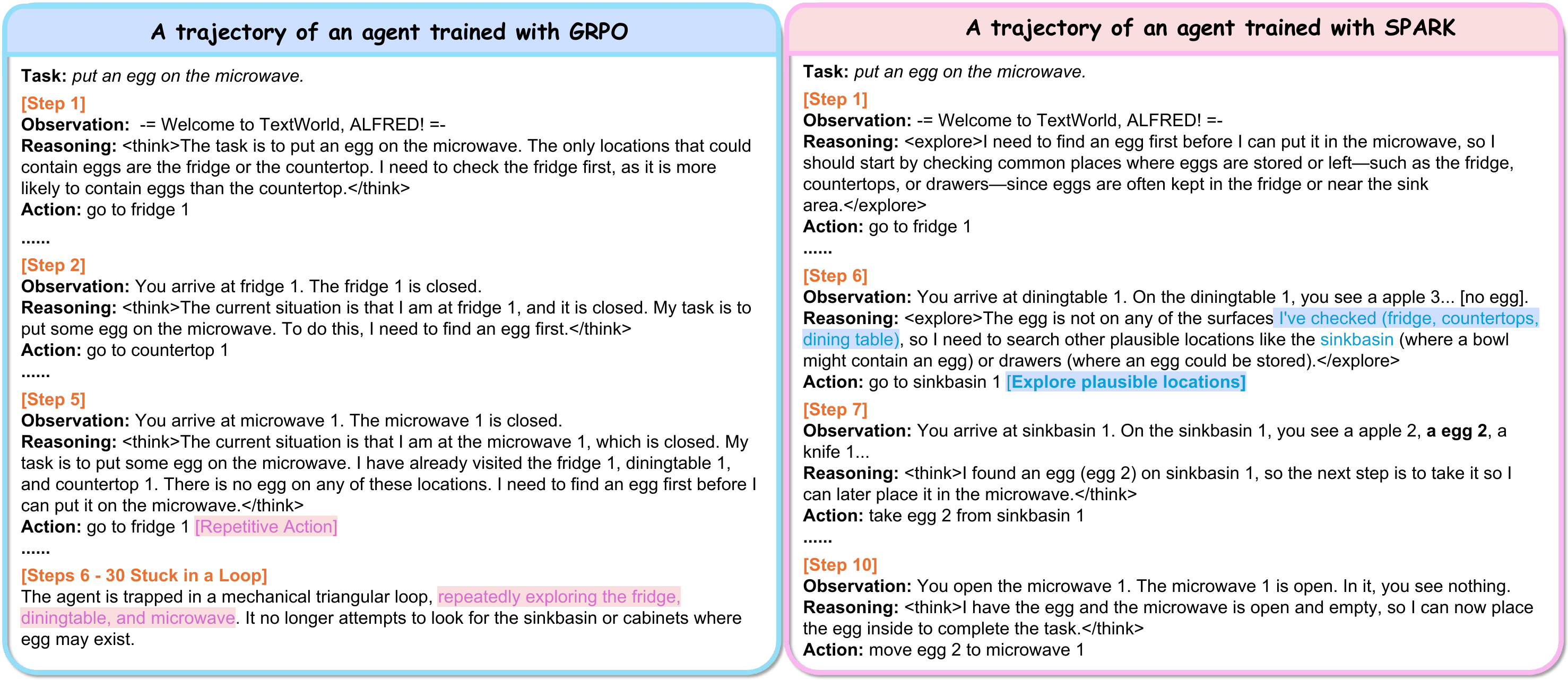}
    \caption{Qualitative comparison on the embodied planning task (ALFWorld): ``put an egg on the microwave''.}
    \label{fig:case_study}
\end{figure*}

\paragraph{Impact of Initial Roots.}
We investigate the trade-off between initial diversity and subsequent exploration depth by varying initial roots ($M$) under fixed budget $N=8$. As shown in Figure~\ref{fig:initial_roots}, performance peaks at $M=4$ (80.7\%), but drops to 67.6\% at $M=2$ and 72.1\% at $M=6$. Too few roots ($M=2$) constrain initial diversity, risking early convergence to local optima despite remaining budget. Conversely, excessive roots ($M\geq6$) exhaust budget at initialization, leaving insufficient resources for later branching. This reveals that the sweet spot $M=4$ balances broad initial coverage with deep adaptive search, splitting budget between diverse starts and strategic exploration.

\begin{table}[h]
\centering
\resizebox{1.0\linewidth}{!}{
\begin{tabular}{lccc}
\toprule
$B$ & ALFWorld L0 & WebShop Succ. & WebShop Score \\
\midrule
2 & \textbf{96.9} & \textbf{75.8} & \textbf{88.8} \\
3 & 94.5 & 73.4 & 87.6 \\
4 & 93.8 & 72.7 & 85.5 \\
\bottomrule
\end{tabular}}
\caption{\textbf{Sensitivity analysis on branching factor $B$} (1.5B backbone). Performance drops as $B$ increases due to premature budget exhaustion.}
\label{tab:branching_factor}
\end{table}

\paragraph{Impact of Branching Factor.}
We conduct an additional sensitivity analysis on the branching factor $B$, which governs the number of parallel continuations generated at each triggered \texttt{<explore>} state.
As shown in Table~\ref{tab:branching_factor}, increasing $B$ from 2 to 4 leads to a consistent performance decline across tasks.
This occurs because a larger $B$ rapidly exhausts the global leaf budget $N$ at a single critical state, leaving insufficient resources for exploration at subsequent pivotal steps later in the trajectory.
This phenomenon mirrors our findings on initial root count $M$ (Section~3.5): just as an excessive $M$ prematurely depletes the budget at initialization, an overly large $B$ over-allocates resources locally, impairing overall trajectory quality.
Thus, $B=2$ strikes the optimal balance between local exploration depth and global budget preservation.

\begin{table}[t]
  \centering
  \resizebox{1.0\linewidth}{!}{
  \begin{tabular}{lcccccccc}
    \toprule
    \multirow{2}{*}{\textbf{Method}} & \multicolumn{3}{c}{\textbf{1.5B}} & & \multicolumn{3}{c}{\textbf{7B}} \\
    \cmidrule(lr){2-4} \cmidrule(lr){6-8}
    & L0 $\downarrow$ & L1 $\downarrow$ & L2 $\downarrow$ & & L0 $\downarrow$ & L1 $\downarrow$ & L2 $\downarrow$ \\
    \midrule
    SFT & 10.7 & 20.2 & 21.4 & & 13.9 & 24.5 & 14.4 \\
    GRPO & 18.4 & 17.6 & 27.1 & & 21.5 & 20.3 & 31.2 \\
    \rowcolor{pink!20}
    Ours & \textbf{10.1} & \textbf{7.0} & \textbf{15.4} & & \textbf{10.0} & \textbf{8.6} & \textbf{14.1} \\
    \bottomrule
  \end{tabular}}
  \caption{\textbf{Repetitive action ratio on ALFWorld} (lower is better). \textsc{Spark} produces fewer repetitive actions on in-domain (L0) and out-of-domain (L1, L2) splits.}
  \label{tab:repetitive_actions}
  \vspace{0.1in}
\end{table}

\paragraph{Exploration Quality: Repetitive Action Ratio.}
Beyond success rates, we analyze exploration quality via repetitive action rates. Table~\ref{tab:repetitive_actions} shows that \textsc{Spark} produces significantly fewer repetitive actions across model scales. On 1.5B's L2 (OOD), \textsc{Spark} achieves 15.4\% repetitive ratio versus GRPO's 27.1\% (43\% reduction). By concentrating search at uncertain states, \textsc{Spark} explores more purposefully, avoiding redundant trial-and-error on routine actions. The gap widens on harder splits (L1, L2), confirming that strategic exploration particularly benefits novel scenarios where indiscriminate search is wasteful.

\begin{table}[t!]
\centering
\resizebox{1.0\linewidth}{!}{
\begin{tabular}{lcccccc}
\toprule
Training Steps & 1 & 30 & 60 & 90 & 120 & 150 \\
\midrule
ALFWorld  & 0.34 & 0.37 & 0.41 & \cellcolor{pink!20}0.48 & 0.47 & 0.29 \\
WebShop   & \cellcolor{pink!20}0.45 & 0.30 & 0.22 & 0.14 & 0.20 & 0.17 \\
\bottomrule
\end{tabular}}
\caption{\textbf{Evolution of \texttt{<explore>} triggering ratio during training}. The mechanism self-regulates, decreasing as the policy matures.}
\label{tab:explore_ratio}
\end{table}

\paragraph{Triggering Frequency Evolution.}
We track the evolution of the exploration ratio (proportion of steps triggering \texttt{<explore>} per trajectory) throughout training.
As shown in Table~\ref{tab:explore_ratio}, the frequency self-regulates rather than saturating or vanishing.
On ALFWorld, the ratio rises initially (Steps 1--90) as the policy explores diverse strategies, then drops at Step 150 (0.29), indicating \emph{policy consolidation}: as the policy improves, the agent autonomously reduces reliance on branching, reserving it for genuinely uncertain states.
On WebShop, the model rapidly identifies that excessive branching is inefficient for linear search-click tasks, converging quickly to a lean exploration rate ($\sim$0.17).
This self-regulating behavior demonstrates that the \texttt{<explore>} mechanism dynamically calibrates to task complexity and policy competency.

\paragraph{Qualitative Analysis.}
Figure~\ref{fig:case_study} illustrates exploration behaviors on an ALFWorld egg-retrieval task. The GRPO-trained agent becomes trapped in a mechanical loop (Steps 6-30), repeatedly checking the same three locations without exploring alternatives like sinkbasin or cabinets. In contrast, \textsc{Spark} exhibits strategic exploration: at Step 6, upon encountering uncertainty about egg location, the agent emits an \texttt{<explore>} signal and reasons about plausible locations (sinkbasin, drawers). This triggers a systematic search across multiple candidates, successfully locating the egg at Step 7 and completing the task by Step 10. This reveals that autonomous identification of critical decision points enables purposeful exploration that avoids redundant loops.

\section{Related Work}
\label{sec:related_work}
\paragraph{Agentic Reinforcement Learning.}
Applying RL to train LLM-based agents has shown promise in reasoning tasks~\citep{shao2024deepseekmath,singh2025agentic,wu2026maestro}, yet long-horizon scenarios remain challenging due to trajectory scarcity with constrained resources. To obtain better training signals, recent work explores process supervision via Process Reward Models~\citep{zhang2025rlvmr,zhang2025process} for dense intermediate feedback, but requires expensive annotations that scale poorly to open-ended environments~\citep{khalifa2025process,shridhar2020alfworld,zhou2024webarena}. Alternative methods attempt to generate high-quality trajectories by expanding exploration budget~\citep{xing2025lookahead,ji2025tree}, but they suffer from indiscriminate resource allocation that treats all steps uniformly. Our method enables agents to autonomously allocate exploration budget, reducing reliance on external supervision while improving sample quality.

\paragraph{Strategic Exploration.}
Effective exploration is critical for discovering optimal policies in vast state-action spaces. Inference-time search methods~\citep{zhang2025survey,wu2024beyond}, such as ToT~\citep{NEURIPS2023_271db992} and RAP~\citep{hao-etal-2023-reasoning}, have demonstrated the value of exploring diverse reasoning paths. Recent efforts aim to internalize this capability into training through tree-structured RL~\citep{treerl,wu2025templaterl}. However, these methods typically apply uniform branching at every step, wasting considerable resources on trivial steps while often under-exploring critical strategic junctures~\citep{ji2025tree}. In contrast, \textsc{Spark} leverages intrinsic decision-making signals to selectively branch only at high-uncertainty states. This design prioritizes sampling quality over exhaustive coverage, achieving superior trajectory quality under budget constraints.

\section{Conclusion}\label{sec:conclusion}
We present \textsc{Spark}, a novel RL framework for long-horizon agentic learning. By enabling strategic exploration through dynamic branching, \textsc{Spark} selectively expands exploration at critical decision points and achieves precise resource allocation. Empirical results confirm that \textsc{Spark} achieves superior success rates with better exploration efficiency, exhibiting robust generalization. Our work highlights strategic exploration's value in advancing capable and efficient agentic systems.

\section*{Limitations}\label{sec:limitation}
Our study is comprehensive, but has certain limitations that we plan to address in future research. In this study, we leverage the agent's intrinsic decision-making signals to identify critical decision points for dynamic branching. While this design minimizes reliance on external supervision and enables autonomous strategic exploration, it may not fully exploit exploration opportunities in extremely low-capability base models where the agent's self-awareness is limited. We believe these are minor issues and we will explore learning-based calibration mechanisms, such as combining internal signals with external feedback to enhance state awareness, thereby further improving the robustness and applicability of strategic branching across diverse model capabilities.

\section*{Acknowledgments}
This work is supported by the National Natural Science Foundation of China (No. U21B2010).

\bibliography{custom}

\appendix

% \clearpage
\newpage
\setcounter{tocdepth}{-1}
\addtocontents{toc}{\protect\setcounter{tocdepth}{2}}

\section*{Appendix of \textsc{Spark}}
This supplementary material provides in-depth insights into our \textsc{Spark} method, covering additional theoretical analysis, additional experimental details, results and analysis. The appendix is organized as follows:

\tableofcontents

\newpage

\section{Theoretical Analysis}\label{sec:appendix_theory}
In this section, we provide \textit{heuristic theoretical justification} for why \textsc{Spark}'s dynamic branching achieves superior results. Our analysis offers conceptual insights into the exploration efficiency under constrained computational budget, serving to motivate our method design and interpret empirical findings. While we present mathematical formulations, we emphasize that these are \textit{informal analyses} rather than rigorous proofs, intended to build intuition for the mechanisms underlying \textsc{Spark}'s performance.

\subsection{Problem Setup and Notations}
Consider a long-horizon agentic task with maximum horizon $K$ (as defined in \S\ref{sec:problem}) and action space $\mathcal{A}$. Let $V_\pi(s)$ denote the value function under policy $\pi$ at state $s$, and define the \textbf{epistemic uncertainty} at state $s$ as:
\[
\begin{aligned}
U(s)
&= \operatorname{Var}_{\pi}\!\left[ Q_\pi(s, a) \right] \\
&= \mathbb{E}_a\!\left[(Q_\pi(s,a) - V_\pi(s))^2\right],
\end{aligned}
\]
where the variance is computed over the action distribution $\pi(\cdot|s)$. Intuitively, $U(s)$ measures the agent's uncertainty regarding the optimal action at state $s$. 

\textbf{Remark on Implicit Estimation:} In our critic-free GRPO framework, while $Q_\pi$ is not explicitly modeled by a value network, the agent performs an \textit{implicit estimation} of this uncertainty through its reasoning trace $z_t$. High $U(s)$ is explicitly manifested as a dedicated \texttt{<explore>} tag in the reasoning trace when the agent recognizes semantic ambiguity or potential missteps.

We partition states along a trajectory into two disjoint sets:
\begin{itemize}
    \item \textit{Critical states} $\mathcal{S}_{\text{crit}} = \{s : U(s) > \tau\}$, where the agent identifies that exploration is valuable.
    \item \textit{Routine states} $\mathcal{S}_{\text{rout}} = \{s : U(s) \leq \tau\}$, where the optimal action is relatively clear and the agent proceeds linearly.
\end{itemize}

Let $K_c = |\mathcal{S}_{\text{crit}} \cap \text{trajectory}|$ denote the number of critical states encountered in a trajectory. For long-horizon agentic tasks, we typically observe $K_c \ll K$ (i.e., most steps are routine navigation).

\subsection{Exploration Efficiency under Budget Constraints}

\noindent\textbf{Analysis 1.} (Exploration efficiency of tree-structured branching) \textit{Consider two exploration strategies with the same rollout budget $N$ (total leaf nodes): (i) \textbf{Independent Sampling}: generate $N$ independent chains, each of length up to $K$; (ii) \textbf{SPARK}: construct a tree with $M$ roots and selective branching at critical states (\textsc{Spark} points). Suppose critical states appear at expected count $K_c$, and let $\Delta_{\text{crit}}$ and $\Delta_{\text{rout}}$ denote the value gain from exploring critical vs. routine states. Our analysis suggests:}
\begin{align*}
\mathbb{E}_{\tau \sim \pi_{\text{\textsc{Spark}}}}[R(\tau)] 
&\gtrsim \mathbb{E}_{\tau \sim \pi_{\text{indep}}}[R(\tau)] \\
&\quad + \Omega\left(\frac{K}{K_c}\right) \cdot (\Delta_{\text{crit}} - \Delta_{\text{rout}}),
\end{align*}
\textit{where the advantage scales with the sparsity ratio $K/K_c$.}

\begin{proof}[Heuristic Argument]
Independent sampling distributes computational effort uniformly across $N \times K$ state-action pairs, resulting in an insufficient exploration depth of $O(1)$ per critical state. \textsc{Spark} instead shares early routine steps (prefixes) across branches, concentrating the budget $B \approx \frac{N}{M \cdot K_c}$ specifically at critical decision points.

Let $f(b)$ denote the value improvement from $b$ exploratory branches. Under the assumption of diminishing returns (concavity of $f$), the total gain can be approximated as:
\[
\begin{aligned}
\text{Gain}_{\text{indep}} 
&\approx N \cdot \left[K_c \cdot \Delta_{\text{crit}} \cdot f(1) \right. \\
&\quad + \left. (K - K_c) \cdot \Delta_{\text{rout}} \cdot f(1)\right] \\
\text{Gain}_{\text{\textsc{Spark}}} 
&\approx M \cdot K_c \cdot \Delta_{\text{crit}} \cdot f\!\left(\frac{N}{M \cdot K_c}\right) \\
&\quad + N \cdot (K - K_c) \cdot \Delta_{\text{rout}} \cdot f(1),
\end{aligned}
\]
When $f$ is concave and $\frac{N}{M \cdot K_c} \gg 1$, the tree structure enables deeper exploration where it matters most, yielding an advantage proportional to the sparsity of critical junctures.
\end{proof}

\subsection{Sample Complexity Advantage}

\noindent\textbf{Analysis 2.} (Reduced sample complexity)
\textit{Let $\epsilon$ denote the target approximation error and
$\rho = K_c / K$ the fraction of critical states along a trajectory.
We argue that \textsc{Spark} can achieve comparable performance using
only an $O(M \cdot \rho)$ fraction of samples relative to uniform
exploration strategies.}

\begin{proof}[Heuristic Argument]
In long-horizon agentic tasks, learning efficiency is primarily governed
by the ability to resolve action preferences at a small number of
\emph{critical decision points}, rather than by uniformly improving
behavior at all steps. Let $T(\epsilon)$ denote the number of comparable
action samples required to identify a near-optimal action at a given
critical state within error $\epsilon$.

Under independent sampling, each trajectory contributes at most one
action sample per critical state. Consequently, after collecting $N$
trajectories, the effective number of samples available for decision
making at each critical state scales as $O(N)$.

In contrast, \textsc{Spark} explicitly branches multiple alternatives
under a shared prefix at each critical state. With a fixed leaf budget
$N$ distributed across $M$ roots and $K_c$ critical states, the effective
number of comparable action samples per critical state scales as
$O\!\left(\frac{N}{M \cdot K_c}\right)$. This concentrated sampling
significantly accelerates the resolution of action uncertainty at
critical decision points.

As a result, to achieve the same decision accuracy $T(\epsilon)$ at each
critical state, \textsc{Spark} requires only an $O(M \cdot \rho)$ fraction
of the total number of trajectories needed by uniform sampling methods,
where $\rho = K_c / K$. Although shared prefixes introduce correlations
between samples, the paired nature of branch comparisons increases the
\textbf{information density} per generated token by avoiding redundant
re-sampling of routine interaction histories.
\end{proof}

\noindent\textbf{Empirical Validation.}
This heuristic scaling behavior is consistent with our empirical
findings in Section~\ref{subsec:efficiency}: \textsc{Spark} matches
RLVMR’s peak performance using only 40\% of the training data, indicating
that comparable performance can be achieved with a constant fraction of
samples. This observation falls within the same order of magnitude as
the estimated critical-state ratio $\rho \approx 0.4$, up to constant
factors.

\begin{table*}[htbp!]
\centering
% \begin{adjustbox}{width=0.85\textwidth}
\resizebox{1.0\linewidth}{!}{
\begin{tabular}{llc}
\toprule
\textbf{Category} & \textbf{Dataset} & \textbf{\#Test Samples} \\
\midrule
\multirow{3}{*}{Embodied Reasoning~\citep{shridhar2020alfworld}}
& ALFWorld-L0 (Seen) & 140 \\
& ALFWorld-L1 (Unseen Instance) & 134 \\
& ALFWorld-L2 (Unseen Category) & 38 \\
\midrule
\multirow{3}{*}{Scientific Reasoning~\citep{scienceworld}}
& ScienceWorld-L0 (Seen) & 1661 \\
& ScienceWorld-L1 (Unseen Instance) & 1684 \\
& ScienceWorld-L2 (Unseen Task) & 549 \\
\midrule
Web Navigation~\citep{yao2022webshop} & WebShop & 500 \\
\bottomrule
\end{tabular}}
% \end{adjustbox}
\caption{Detailed information on the agentic benchmarks and test set split sizes.}
\label{table:dataset-details}
\end{table*}

\subsection{Summary}
Our heuristic theoretical analysis establishes three key insights:
\begin{enumerate}
    \item \textbf{Exploration Efficiency} (Analysis 1): Tree-structured branching concentrates budget at critical states, suggesting an $\Omega(K/K_c)$ advantage over independent sampling when $K_c \ll K$.
    \item \textbf{Sample Complexity} (Analysis 2): \textsc{Spark} may require approximately an $M \cdot \rho$ fraction of samples ($\rho = K_c/K$) to match uniform methods' performance under favorable conditions.
    % \item \textbf{Generalization} (Analysis 3): The meta-strategy $\phi$ has lower complexity than the full policy $\pi$, suggesting a smaller generalization gap across tasks.
\end{enumerate}

These analyses provide conceptual justification and intuition for \textsc{Spark}'s empirical success: superior performance (\textit{Sec.}~\ref{subsec:main_results}), and sample efficiency (\textit{Sec.}~\ref{subsec:efficiency}). While not rigorous proofs, they offer valuable theoretical perspective on the mechanisms underlying our method's effectiveness.

\section{Additional Experimental Details}\label{sec:experimental_details}
\subsection{Datasets}\label{subsec:datasets}
The evaluation benchmarks used in this paper are detailed in Table \ref{table:dataset-details}. Following prior work~\citep{feng2025groupingroup, zhang2025rlvmr}, we evaluate our method on three long-horizon agentic domains. To rigorously test the generalization of our \textsc{Spark} paradigm, we categorize tasks into three levels: \textbf{L0} (seen categories/instances), \textbf{L1} (seen categories but unseen instances), and \textbf{L2} (unseen categories and instances). Details are listed as follows:

\begin{itemize}[leftmargin=1.36em]
    \item \textbf{ALFWorld~\citep{shridhar2020alfworld}}: This benchmark aligns TextWorld with the ALFRED environment, requiring agents to execute household tasks (e.g., \textit{Look}, \textit{Clean}, \textit{Pick2}) in a text-based interactive world. It poses challenges in high-level goal decomposition and low-level action execution. We specifically focus on the \textit{Unseen} split (L2) to evaluate the agent's ability to generalize its learned exploration strategies to completely novel domestic environments.

    \item \textbf{ScienceWorld~\citep{scienceworld}}: ScienceWorld is a complex text-based environment simulating 30 distinct scientific tasks across 10 virtual environments. It requires the agent to apply the scientific method (e.g., conducting experiments, measuring substances) with deep horizons often exceeding 30+ steps. Due to its enormous state space and strict logic requirements, it serves as a robust testbed for our strategic exploration via dynamic branching mechanism.

    \item \textbf{WebShop~\citep{yao2022webshop}}: This dataset provides a realistic e-commerce simulation containing over 1.1 million products. Agents must navigate via search and button clicks to purchase items that match specific user requirements. Evaluation is conducted on 500 test queries, where we report both the \textit{task completion score} and \textit{success rate}. WebShop tests the model's ability to perform open-ended exploration and attribute-matching in large-scale, semi-structured observation spaces.
\end{itemize}

\begin{figure*}[t]
  \centering
  \includegraphics[width=0.96\linewidth]{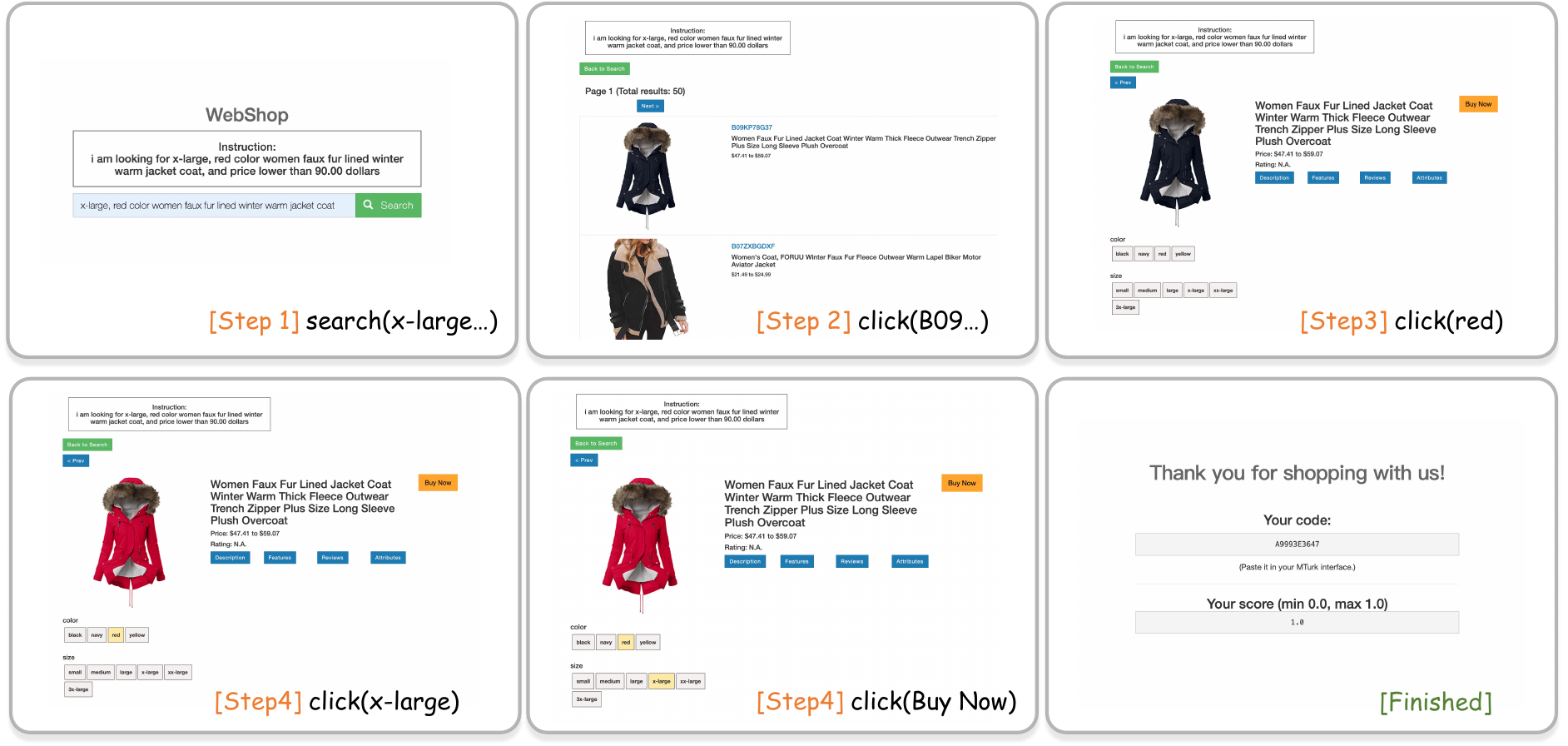}
  \caption{\textbf{Illustration of Webshop Environment.} We illustrate the agent's execution path: first searching for items, then clicking on a suitable product, selecting its color and size, and finally completing the purchase.}
  \label{fig:webshop_show}
  \vspace{-0.1in}
\end{figure*}

\paragraph{Task Illustration.}
We illustrate interactions between agent and Webshop environment in Figure \ref{fig:webshop_show}. 

\subsection{Baselines}\label{subsec:baselines}
We evaluate \textsc{Spark} against three representative classes of baselines, which are briefly described as follows:

\paragraph{\textbf{(1) Closed-source LLMs}: }
\begin{itemize}
    \item \textbf{GPT-4o}~\citep{gpt4o}: A natively multimodal large language model that can process and generate text, images, and audio within a single architecture, enabling real-time interaction, improved cross-modal reasoning, and low-latency conversational performance.

    \item \textbf{GPT-5-mini}~\citep{gpt5}: A compact and resource-efficient variant of the GPT-5 family, designed to deliver reliable language understanding and reasoning while minimizing computational cost.

    \item \textbf{GPT-5}~\citep{gpt5}: A next-generation flagship model focused on deep reasoning, long-context understanding, and complex problem solving across domains such as science, coding, and analysis, with enhanced alignment and safety mechanisms.

    \item \textbf{Gemini-2.5-Pro}~\citep{comanici2025gemini}: An advanced multimodal model developed by Google DeepMind, optimized for strong logical reasoning, mathematical problem solving, and code generation, with robust long-context processing capabilities.
\end{itemize}

\paragraph{\textbf{(2) Prompting Methods}: }
\begin{itemize}
    \item \textbf{ReAct}~\citep{yao2022react}: 
    A synergistic framework that interleaves reasoning traces with task-specific actions, enabling LLMs to interact with external environments in real time.
    ReAct has been widely adopted in agentic tasks and serves as a foundational paradigm shared by many methods.
\end{itemize}
\paragraph{\textbf{(3) RL Training Methods}: }
\begin{itemize}
    \item \textbf{ETO}~\citep{song2024trial}:  An exploration-centric paradigm that enhances model policies through a trial-and-error mechanism. By constructing contrastive success-failure pairs from collected trajectories, it enables LLMs to iteratively refine their decision-making logic and learn from past procedural errors.
    \item \textbf{GRPO}~\citep{shao2024deepseekmath}: A critic-free reinforcement learning framework that estimates advantages through relative rewards within sampled groups. By eliminating the need for a separate value model, it significantly optimizes computational efficiency while maintaining robust performance in reasoning-intensive tasks.
    \item \textbf{GiGPO}~\citep{feng2025groupingroup}: 
    A hierarchical policy optimization method that introduces a group-in-group structure for fine-grained credit assignment. By clustering anchor states across trajectories, it enables precise optimization of long-horizon tasks without the dependency on dense external reward models.

    \item \textbf{RLVMR}~\citep{zhang2025rlvmr}: 
    A meta-reasoning approach that integrates verifiable cognitive steps into the reinforcement learning loop. By rewarding explicit planning and reflection behaviors via rule-based verification, it enhances the transparency and generalizability of agents in complex problem-solving.
\end{itemize}

\vspace{-0.1in}

\begin{figure*}[t]
  \centering
  \includegraphics[width=0.92\linewidth]{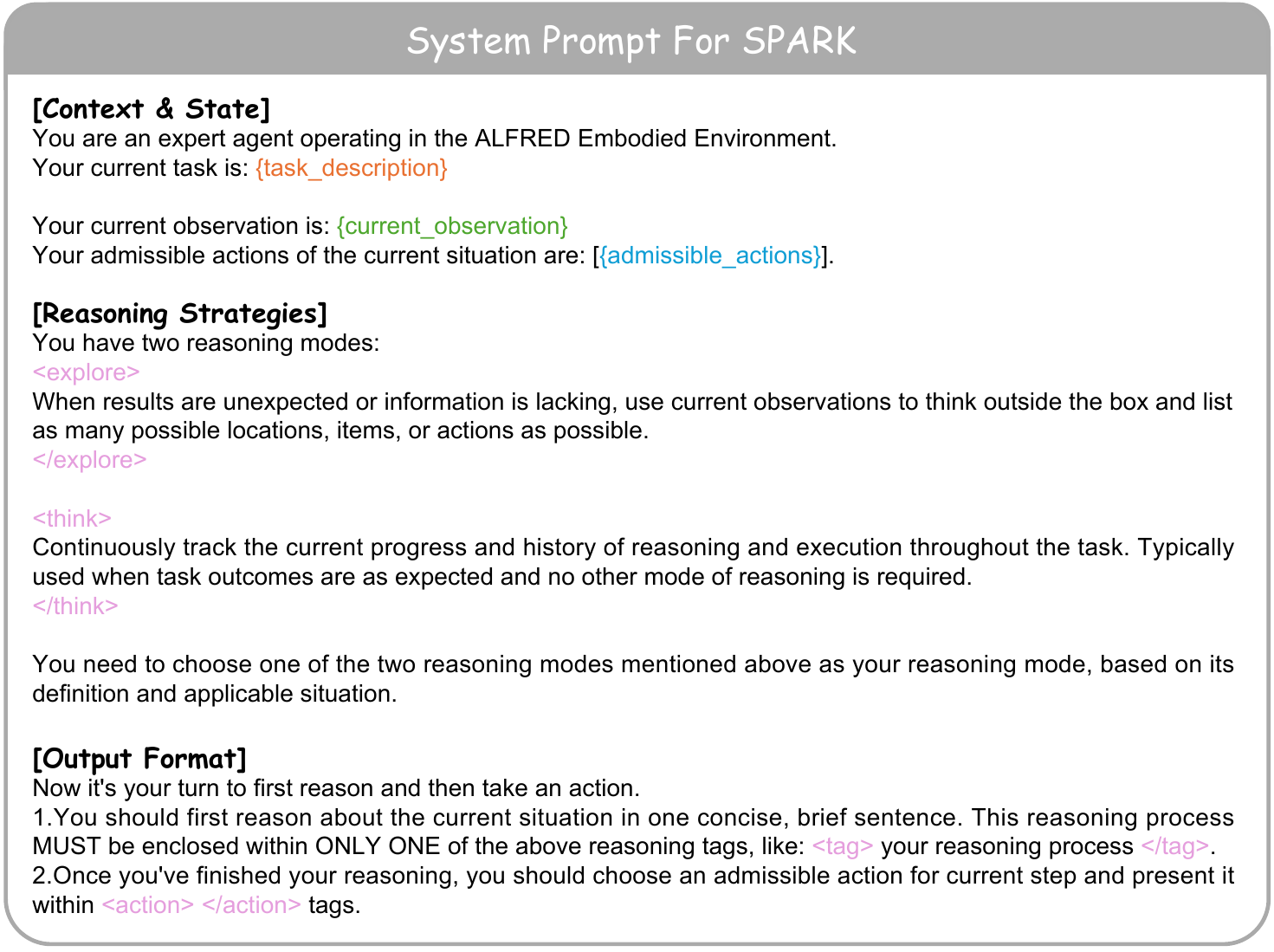}
  \caption{System prompt for RL in \textsc{Spark}.}
  \label{fig:system_prompt}
  \vspace{-0.1in}
\end{figure*}

\subsection{Evaluation Details}\label{subsec:evaluation_details}
In our evaluation, we consistently adopt Success Rate (SR.) as the primary performance metric. The success rate measures whether an agent successfully completes a given task.
For the WebShop environment, in addition to success rate, we also report the task completion score (Score), which reflects the degree of task completion. Specifically, even when a task is not fully completed, the agent can still receive partial credit based on the completion of intermediate sub-goals.

For all evaluation experiments, we set the test batch size to 128, and fix the random seed to 0 to ensure reproducibility and fair method comparison.

\vspace{-0.1in}
\subsection{Implementation Details}\label{subsec:implementation}
\paragraph{Details for SFT.}
Our SFT dataset consists of 300 successful trajectories designed to initialize the model's capability to generate intrinsic exploration signals.
The dataset is a hybrid constructed from two distinct sources: 90\% via a \emph{retro-annotation strategy} and 10\% via \emph{real interactions}.

\begin{table}[ht]
\centering
\resizebox{1.0\linewidth}{!}{
\begin{tabular}{lcccccc}
\toprule
Step Interval & 1--5 & 6--10 & 11--15 & 16--20 & 21--25 & 26--30 \\
\midrule
Ratio & 22.97\% & 10.43\% & 14.20\% & 7.69\% & 3.57\% & 0.00\% \\
\bottomrule
\end{tabular}}
\caption{\textbf{Distribution of \texttt{<explore>} tags across step intervals in ALFWorld SFT data.} Tags are concentrated in early steps, reflecting temporal exploration necessity.}
\label{tab:explore_distribution}
\vspace{-0.1in}
\end{table}

\vspace{0.1in}
\noindent\textbf{(1) Retro-annotation (90\%).}
The ``golden paths'' are sourced from the datasets used by the RLVMR baseline~\cite{zhang2025rlvmr}, which contain only observation-action (obs, action) pairs without intermediate reasoning steps.
We employ a teacher model (Kimi-K2) to perform retro-annotation: it reviews these static sequences and synthesizes missing reasoning traces, selectively inserting \texttt{<explore>} tags at steps where epistemic uncertainty logically arises.

\vspace{0.1in}
\noindent\textbf{(2) Real interactions (10\%).}
These are entirely new successful trajectories generated organically from scratch by deploying the teacher model to interact directly with the environments.
When the teacher model successfully completes a task, the full trajectory---including spontaneous reasoning steps with \texttt{<explore>} and \texttt{<think>} tags---is recorded and added to the SFT dataset.

\vspace{0.1in}
\noindent\textbf{Tag Placement Robustness.}
We find the method robust to variations in tag quality and placement.
Replacing \texttt{<explore>} with a semantically equivalent alternative (\texttt{<exploring>}) yields no meaningful performance degradation.
Furthermore, as shown in Table~\ref{tab:explore_distribution}, the \texttt{<explore>} tags are concentrated in early trajectory steps to address initial ambiguity, and decrease as the task progresses, confirming that exploration signals are distributed based on temporal necessity rather than random placement.

\begin{table}[ht!]
\centering
\resizebox{1.0\linewidth}{!}{
\begin{tabular}{lcccc}
\toprule
\multirow{2}{*}{\textbf{SFT Data Source}} & \multicolumn{2}{c}{\textbf{ALFWorld}} & \multicolumn{2}{c}{\textbf{ScienceWorld}} \\
\cmidrule(lr){2-3} \cmidrule(lr){4-5}
 & L0 & L2 & L0 & L2 \\
\midrule
\rowcolor{pink!20}
Kimi-K2 (Default) & \textbf{96.9} & \textbf{80.5} & \textbf{69.5} & \textbf{49.2} \\
Gemini-2.5-Flash  & 95.8 & 79.5 & 67.6 & 48.1 \\
Qwen2.5-32B       & 95.6 & 78.9 & 67.7 & 47.8 \\
\bottomrule
\end{tabular}}
\caption{\textbf{Effect of SFT teacher model on final performance} (1.5B backbone). Performance differences are marginal across teacher models, confirming that SFT serves as format alignment.}
\label{tab:sft_teacher}
\end{table}

\vspace{0.1in}
\noindent\textbf{Effect of SFT Phase.}
To further validate that the cold-start SFT phase functions as \emph{format alignment} rather than distilling strategic priors from a specific frontier model, we compare the final performance of SPARK when the SFT trajectories are synthesized by different teacher models. As shown in Table~\ref{tab:sft_teacher}, substituting Kimi-K2 with Gemini-2.5-Flash or Qwen2.5-32B yields only marginal differences in downstream performance.
This demonstrates that SPARK does not heavily rely on the advanced capabilities of any particular teacher model; the true strategy optimization occurs during the RL phase through environment interaction.

\paragraph{Details for RL.}
Response lengths are set to 512 tokens. Each episode is capped at 30 environment steps. We employ a learning rate of $1\times10^{-6}$ for the actor. A rule-based reward system is adopted, assigning a reward of 10 for task success and 0 for failure. To encourage adherence to the environment's action space, a penalty of -0.1 is applied for any invalid actions generated by the agent. For all group-based RL methods, the total budget (i.e. group size $N$) is set to 8, with a batch size of 16. We use a temperature of 0.4 in rollout phase. The KL-divergence loss coefficient is fixed at 0.01. To prevent context window overflow, we restrict the conversation history to a maximum length of 5. Specifically for \textsc{Spark}, the number of initial roots is set to 4 and branching factor is set to 2 by default. For ALFWorld, ScienceWorld, and WebShop, the maximum prompt lengths are set to 2,048, 4,096, and 6,000 tokens, respectively. The system prompt is shown in Figure~\ref{fig:system_prompt}.

\vspace{-0.1in}
\paragraph{Computing Details.} All experiments are conducted on 4 NVIDIA A100-80GB GPUs.

\begin{table}[htbp]
\centering
\resizebox{1.0\linewidth}{!}{
\begin{tabular}{lccccc}
\toprule
\textbf{Method} & \textbf{20\%} & \textbf{40\%} & \textbf{60\%} & \textbf{80\%} & \textbf{100\%} \\
\midrule
GRPO & 22.7 & 37.5 & 46.1 & 61.8 & 76.6 \\
GiGPO & 22.7 & 41.4 & 68.8 & 76.6 & 86.7 \\
RLVMR & 68.8 & 71.1 & 86.7 & 86.7 & 89.1 \\
\rowcolor{pink!20}
Ours & \textbf{84.4} & \textbf{89.1} & \textbf{91.4} & \textbf{96.6} & \textbf{96.9} \\
\bottomrule
\end{tabular}}
\caption{\textbf{Sample efficiency comparison on ALFWorld L0}. We report success rates (\%) on Qwen2.5-1.5B under varying training data percentages. \textsc{Spark} consistently outperforms representative RL baselines.}
\label{tab:sample_efficiency}
\vspace{-0.1in}
\end{table}

\section{Supplementary Results}\label{sec:results}
\subsection{Sample Efficiency}\label{subsec:sample_efficiency}
We provide detailed sample efficiency results in Table~\ref{tab:sample_efficiency}. The results demonstrate that \textsc{Spark} exhibits superior sample-efficient scaling properties: with only 20\% of the training data, our method already achieves a success rate of 84.4\%, which significantly surpasses the performance of the GRPO baseline trained on the full (100\%) dataset (76.6\%). Furthermore, \textsc{Spark} matches the strong RLVMR baseline's peak performance using only 40\% total samples, highlighting the efficacy of our strategic exploration mechanism in extracting higher information density from limited environment interactions. Similar results are also observed in Table~\ref{tab:sample_efficiency_sci}.

\subsection{Inference Scalability}\label{subsec:inference_scalability}
To further investigate the performance potential of our model under extended computational budgets, we evaluate the inference scalability of \textsc{Spark} compared to other baselines. For each task across the three environments, we generate $n=32$ candidate trajectories and report the success rate using the $\text{Pass@}k$ metric with $k=16$. Following standard definition \citep{chen2021evaluatinglargelanguagemodels}, the $\text{Pass@}k$ is:
$$\text{Pass@}k := \mathbb{E}_{\text{Problems}} \left[ 1 - \frac{\binom{n-c}{k}}{\binom{n}{k}} \right]$$
where $n$ is the total number of sampled trajectories, $c$ is the number of successful trajectories among $n$ samples, and $k$ is the evaluation threshold. As shown in Table~\ref{tab:inference_scaling}, our method consistently outperforms all competitive baselines across all benchmarks. Notably, in the complex ScienceWorld environment, \textsc{Spark} achieves a success rate of 94.9\%, surpassing the strongest baseline by 30.1\%. Results show that by prioritizing sampling quality over blind coverage, \textsc{Spark} significantly enhances model ability to solve long-horizon tasks when more inference-time budget is allocated.

\begin{table}[t]
\centering
\resizebox{1.0\linewidth}{!}{
\begin{tabular}{lccccc}
\toprule
\textbf{Method} & \textbf{20\%} & \textbf{40\%} & \textbf{60\%} & \textbf{80\%} & \textbf{100\%} \\
\midrule
GRPO & 0.0 & 2.3 & 3.1 & 9.0 & 21.1 \\
GiGPO & 1.6 & 1.9 & 10.2 & 21.9 & 25.8 \\
RLVMR & 30.4 & 39.6 & 43.4 & 47.6 & 46.9 \\
\rowcolor{pink!20}
Ours & \textbf{36.6} & \textbf{47.3} & \textbf{53.3} & \textbf{60.7} & \textbf{69.5} \\
\bottomrule
\end{tabular}}
\caption{\textbf{Sample efficiency comparison on ScienceWorld L0}. We report success rates (\%) on Qwen2.5-1.5B under varying training data percentages. \textsc{Spark} consistently outperforms RL baselines.}
\label{tab:sample_efficiency_sci}
\vspace{-0.05in}
\end{table}

\begin{table}[t]
    \centering
    \resizebox{1.0\linewidth}{!}{
        \begin{tabular}{lccc}
        \toprule
        Method & ALFWorld$\uparrow$ & ScienceWorld$\uparrow$ & WebShop$\uparrow$ \\
        \midrule
        ReAct   & 19.8 & 4.5  & 18.6 \\
        SFT     & 93.7 & 64.8 & 59.7 \\
        GRPO    & 90.9 & 45.5 & 78.8 \\
        GiGPO   & 97.0 & 53.1 & 74.0 \\
        RLVMR   & 99.9 & 64.8 & 75.4 \\
        \rowcolor{pink!20}
        Ours & \textbf{100.0} & \textbf{94.9} & \textbf{82.0} \\
        \bottomrule
        \end{tabular}}
    \caption{\textbf{Inference Scalability Analysis.} We report the $\text{Pass@}16$ success rates (\%) on ALFWorld, ScienceWorld, and WebShop using Qwen2.5-1.5B. Results show that \textsc{Spark} significantly outperforms competitive baselines across all environments.}
    \label{tab:inference_scaling}
    \vspace{-0.2in}
\end{table}

\begin{figure*}[t]
  \centering
  \includegraphics[width=0.96\linewidth]{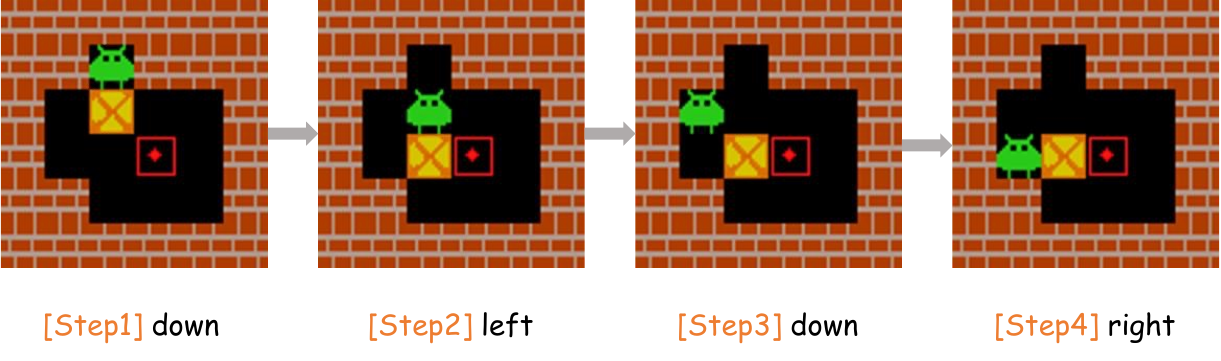}
  \caption{\textbf{Illustration of Sokoban Environment.} We illustrate the agent's execution path: down, left, down, right and finally reach the goal.}
  \label{fig:sokoban_show}
  % \vspace{-0.1in}
\end{figure*}

\subsection{Multimodal Extension}\label{subsec:multimodal_extension}
To validate the generalizability of our framework beyond text-only tasks, we extend our evaluation to the multimodal setting using Qwen2.5-VL-3B-Instruct~\citep{bai2025qwen2}. We employ two distinct visual benchmarks: Sokoban~\citep{SchraderSokoban2018} ($6\times6$ size), a classic grid-based puzzle that necessitates spatial reasoning and long-term planning to push boxes onto designated targets, and EZPoints in Gym Cards~\citep{NEURIPS2024_c848b7d3}, a visual reasoning task where the agent must perceive playing cards and formulate arithmetic steps to reach a target value of 12. As shown in Table \ref{tab:multimodal_extension}, \textsc{Spark} achieves substantial improvements across all baseline approaches. Compared to GRPO, our method demonstrates 11.3 points gain in average success rate (88.3\% vs. 77.0\%), with particularly notable improvements on Sokoban (17.3 points) and consistent gains on EZPoints (5.2 points). More remarkably, our approach surpasses GPT-5 by 16.4 points (88.3\% vs. 71.9\%) and dramatically outperforms the ReAct prompting strategy by 80.9 points (88.3\% vs. 7.4\%). These results highlight the broad applicability of our framework, demonstrating that our strategic exploration approach is modality-agnostic. As shown in Figure~\ref{fig:sokoban_show}, we provide a trace visualization on Sokoban.
\begin{table}[t]
    \centering
    \resizebox{\linewidth}{!}{
        \begin{tabular}{lccc}
        \toprule
        Method & Sokoban[6$\times$6]$\uparrow$ & EZPoints$\uparrow$ & Average$\uparrow$ \\
        \midrule
        GPT-5   & 45.3 & 98.4 & 71.9\\
        ReAct   & 11.7 & 3.1 &  7.4\\
        GRPO    & 67.1 & 86.9 & 77.0\\
        \rowcolor[RGB]{236,244,252}
        Ours & \textbf{84.4} & \textbf{92.1} & \textbf{88.3}\\
        \rowcolor{pink!20}
        $\bigtriangleup$ vs. GRPO& +17.3 & +5.2 & +11.3\\
        \bottomrule
        \end{tabular}}
    \caption{\textbf{Multimodal Extension Analysis.} We present success rates (\%) on two multimodal visual benchmarks, Sokoban[6$\times$6] and EZPoints, using Qwen2.5-VL-3B-Instruct as the underlying backbone model.}
    \label{tab:multimodal_extension}
\end{table}

\begin{table}[t]
    \centering
    % \resizebox{0.8\linewidth}{!}{
        \begin{tabular}{lccc}
        \toprule
        Budget ($N$) & 4 & 8 & 16 \\
        \midrule
        L0 & 89.1 & \cellcolor{pink!20}96.9 & \textbf{97.7}\\
        L1 & 81.3 & \cellcolor{pink!20}\textbf{93.8} & 91.4\\
        \bottomrule
        \end{tabular}
        % }
    \caption{\textbf{Sensitivity Analysis on Total Budget($N$).} We investigate how the model performance varies with different budget constraints ($N \in \{4, 8, 16\}$) under ALFWorld L0 and L1, based on 1.5B backbone.}
    \label{tab:sensitivity_analysis}
\end{table}

\subsection{Sensitivity Analysis on Total Budget}\label{appendix_subsec_sensitivity}
As shown in Table \ref{tab:sensitivity_analysis}, we conduct a sensitivity analysis on the total budget $N$. Increasing $N$ from 4 to 8 yields substantial gains (+7.8\% on L0, +12.5\% on L1), demonstrating that \textsc{Spark} effectively leverages additional computational resources for enhanced exploration. However, further scaling to $N=16$ shows diminishing returns, with performance plateauing. This pattern suggests that $N=8$ strikes an optimal balance between exploration breadth and efficiency, beyond which marginal benefits diminish. These results confirm that our method achieves strong cost-effectiveness without requiring excessive budget allocation.

\begin{table}[h]
\centering
\resizebox{1.0\linewidth}{!}{
\begin{tabular}{lccc}
\toprule
Method & ALFWorld & ScienceWorld & WebShop \\
\midrule
baseline-inference & 96.9 & 69.5 & 75.8 \\
spark-inference    & \cellcolor{pink!20}\textbf{98.4} & \cellcolor{pink!20}\textbf{71.9} & \cellcolor{pink!20}\textbf{78.8} \\
\bottomrule
\end{tabular}}
\caption{\textbf{Test-time branching results.} Applying \texttt{<explore>}-conditioned branching at inference yields consistent improvements, indicating potential for Test-Time Scaling.}
\label{tab:test_time}
\end{table}

\subsection{Test-Time Branching}\label{appendix:test_time}
The \texttt{<explore>} signal, originally designed to govern training-time rollout topology, also serves as a principled indicator of decision uncertainty at inference time.
To assess its test-time utility, we implement a \emph{spark-inference} strategy that applies conditional branching when the model emits \texttt{<explore>} during standard autoregressive generation, selecting the best continuation via majority voting.

As shown in Table~\ref{tab:test_time}, spark-inference yields consistent gains across all benchmarks, demonstrating that the learned \texttt{<explore>} signal retains meaningful utility beyond the training phase and aligns with the emerging paradigm of Test-Time Scaling~\cite{zhang2025survey}.

\subsection{Statistical Analysis}\label{subsec:statistical_analysis}
To statistically evaluate the significance of performance differences between \textsc{Spark} and baseline methods, we use the nonparametric Wilcoxon signed-rank test~\citep{wilcoxon1992individual}. This statistical test is specifically designed to compare paired observations when data may not follow a normal distribution, making it particularly suitable for performance analysis across methods. The Wilcoxon signed-rank test evaluates whether there is a significant difference between paired observations through the following procedure:
\begin{enumerate}
\item \textbf{Calculate differences:} For each benchmark pair, compute the difference $D_i = X_i - Y_i$ where $X_i$ represents \textsc{Spark} performance and $Y_i$ represents baseline performance.
\item \textbf{Rank differences:} Take absolute values $|D_i|$ and rank them from smallest to largest as $R_i$, with average ranks assigned for ties.
\item \textbf{Assign signs to ranks:} For each difference $D_i$, assign its sign to the corresponding rank: $R'_i = \text{sign}(D_i) \cdot R_i$.
\item \textbf{Calculate rank sums:} Compute positive and negative rank sums: $W^+ = \sum_{D_i > 0} R'_i$ and $W^- = \sum_{D_i < 0} R'_i$.
\item \textbf{Determine test statistic:} The test statistic is $W = \min(W^+, W^-)$.
\item \textbf{Calculate p-value:} Derive the p-value from the distribution of test statistic $W$.
\end{enumerate}

We conducted the Wilcoxon test by pairing \textsc{Spark} results against GRPO across all benchmarks and model scales (1.5B and 7B). The null hypothesis is $H_0$: no significant difference between methods, while the alternative hypothesis is $H_1$: significant difference exists. \textbf{The analysis yields a p-value of 9.7e-4} (calculated based on the consistent dominance across multiple task domains shown in Table 1). At a significance level of $\alpha = 0.05$, we \textbf{reject the null hypothesis} ($p < \alpha$), providing compelling statistical evidence that \textsc{Spark} possesses a significant advantage over standard GRPO. This confirms that autonomous strategic exploration facilitates a robust and reproducible improvement in agentic reasoning.

\section{Case Study}\label{sec:case_study}
\begin{figure*}[p]
    \includegraphics[width=\linewidth]{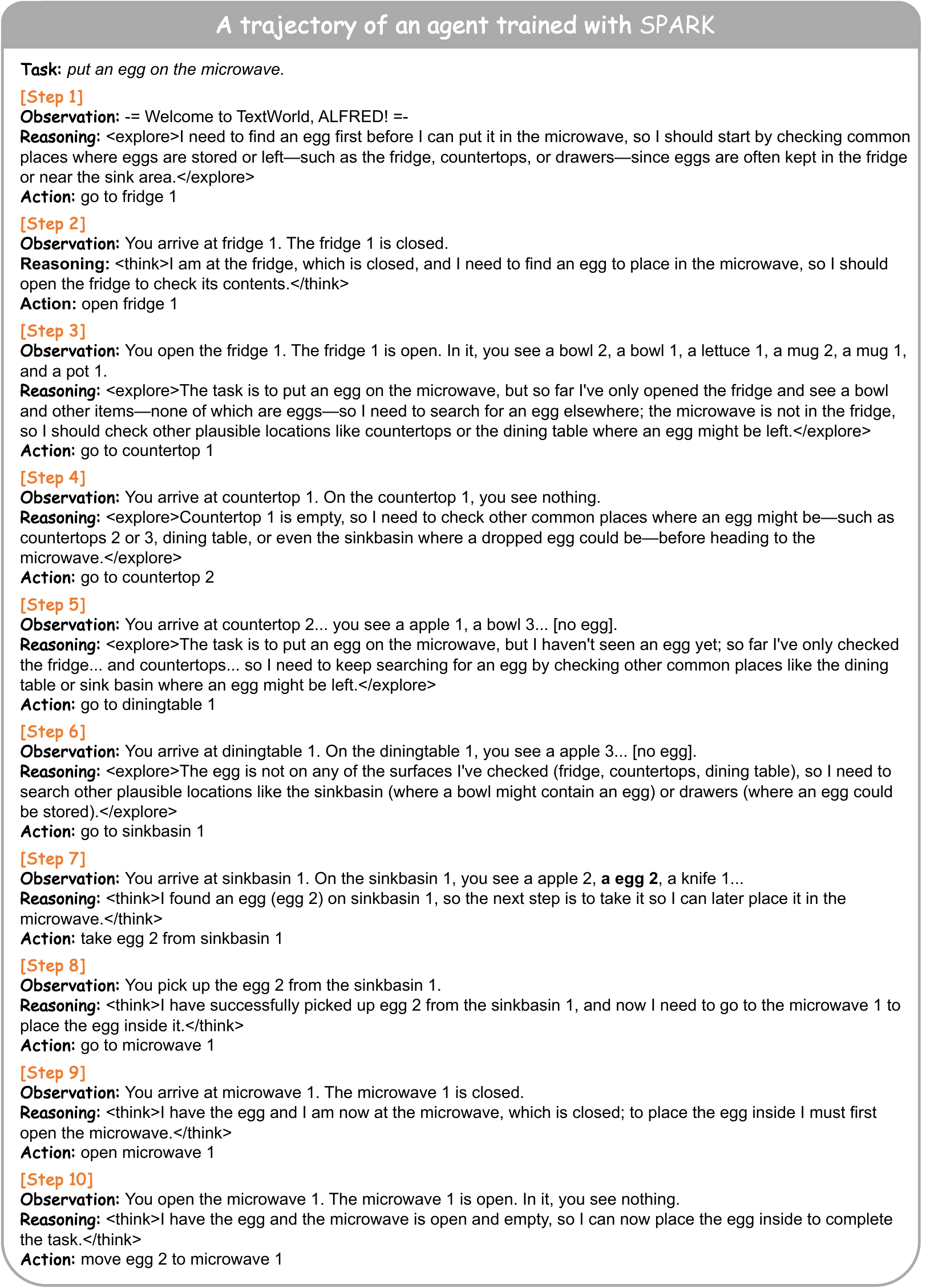}
    \caption{A full trajectory of \textsc{Spark} in ALFWorld (Example 1).}
    \label{fig:fullcase_alfworld}
\end{figure*}

\begin{figure*}[p]
    \includegraphics[width=\linewidth]{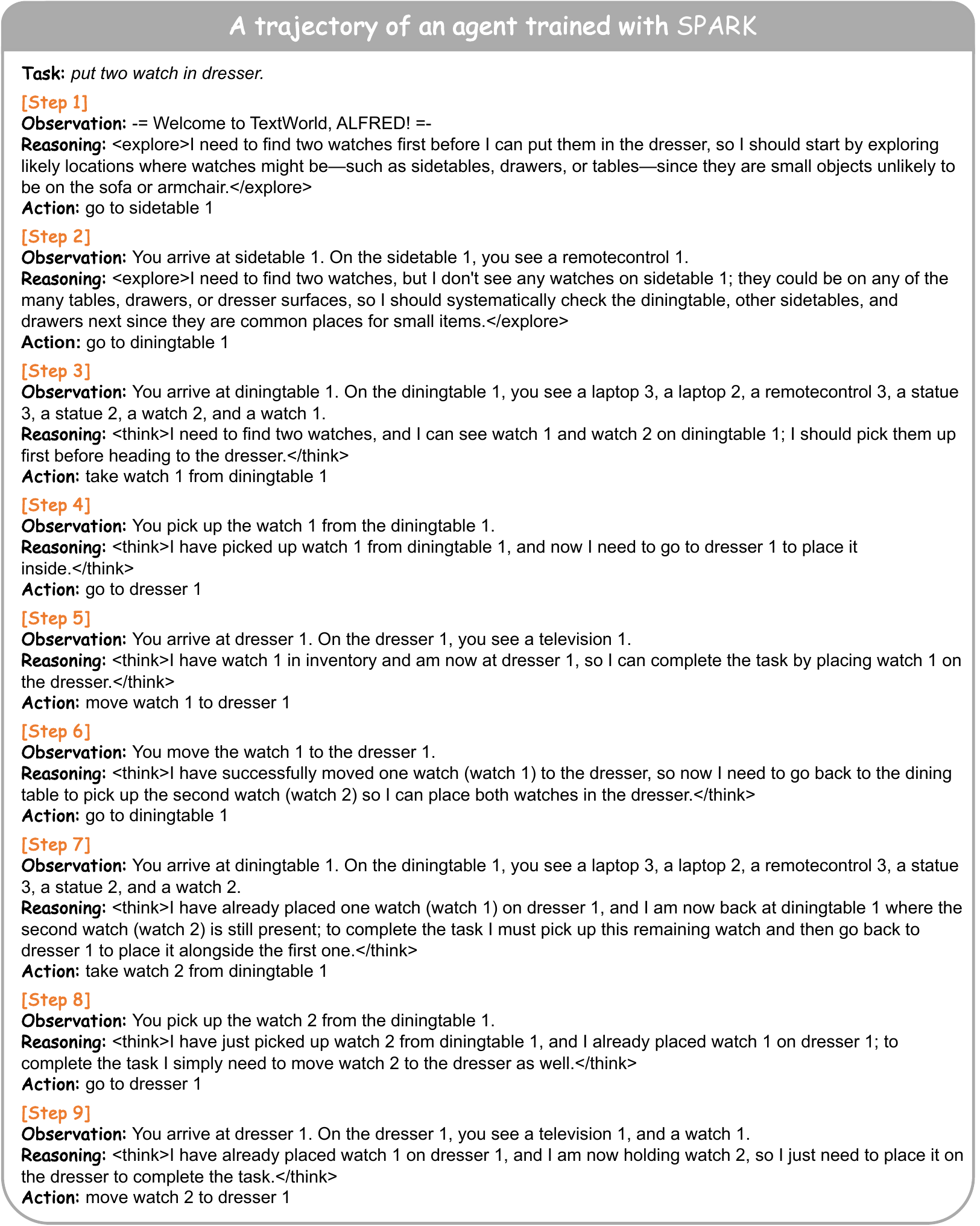}
    \caption{A full trajectory of \textsc{Spark} in ALFWorld (Example 2).}
    \label{fig:fullcase_alfworld2}
\end{figure*}

\begin{figure*}[p]
    \includegraphics[width=\linewidth]{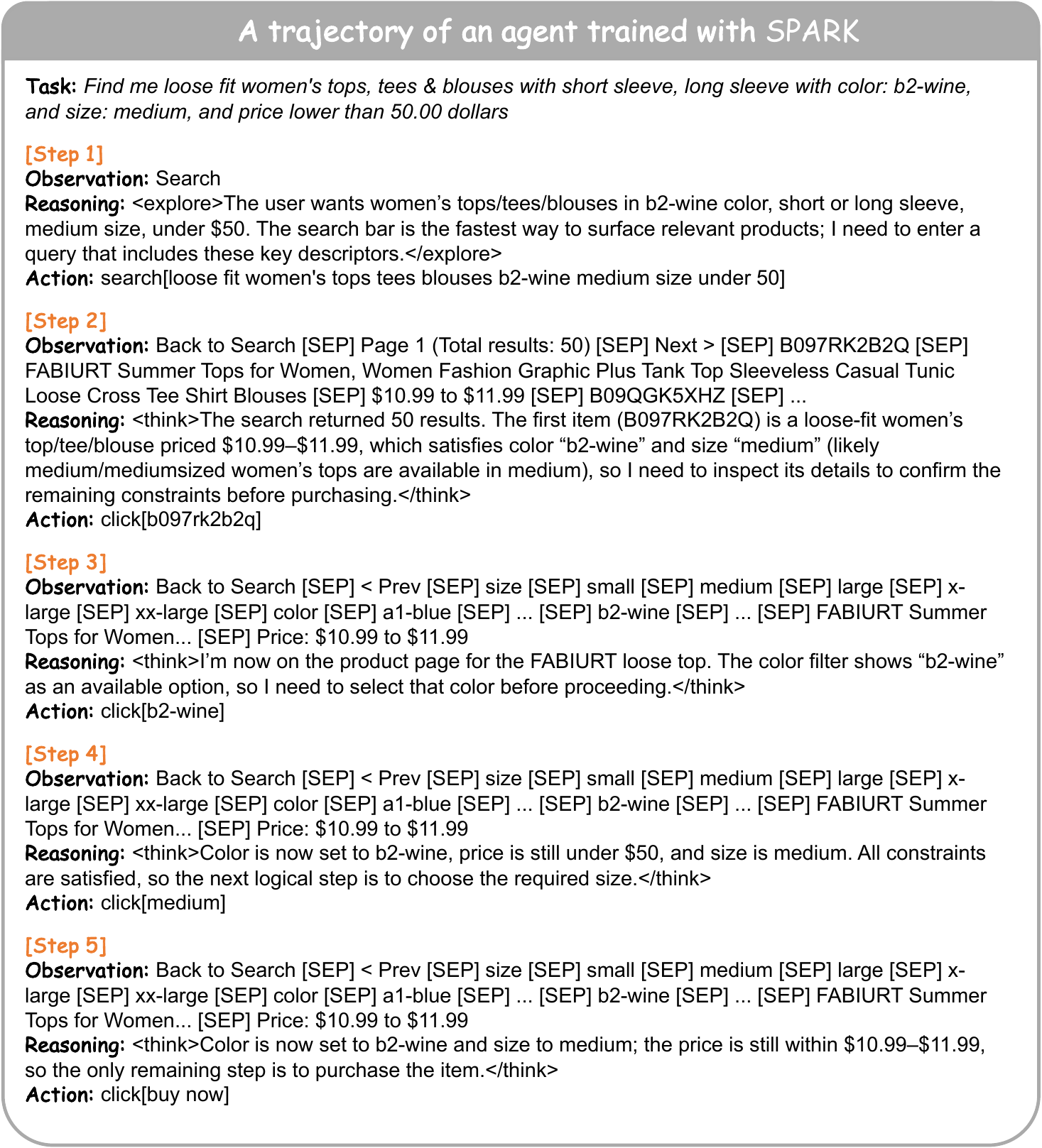}
    \caption{A full trajectory of \textsc{Spark} in WebShop (Example 3).}
    \label{fig:fullcase_webshop}
\end{figure*}

\begin{figure*}[p]
    \includegraphics[width=\linewidth]{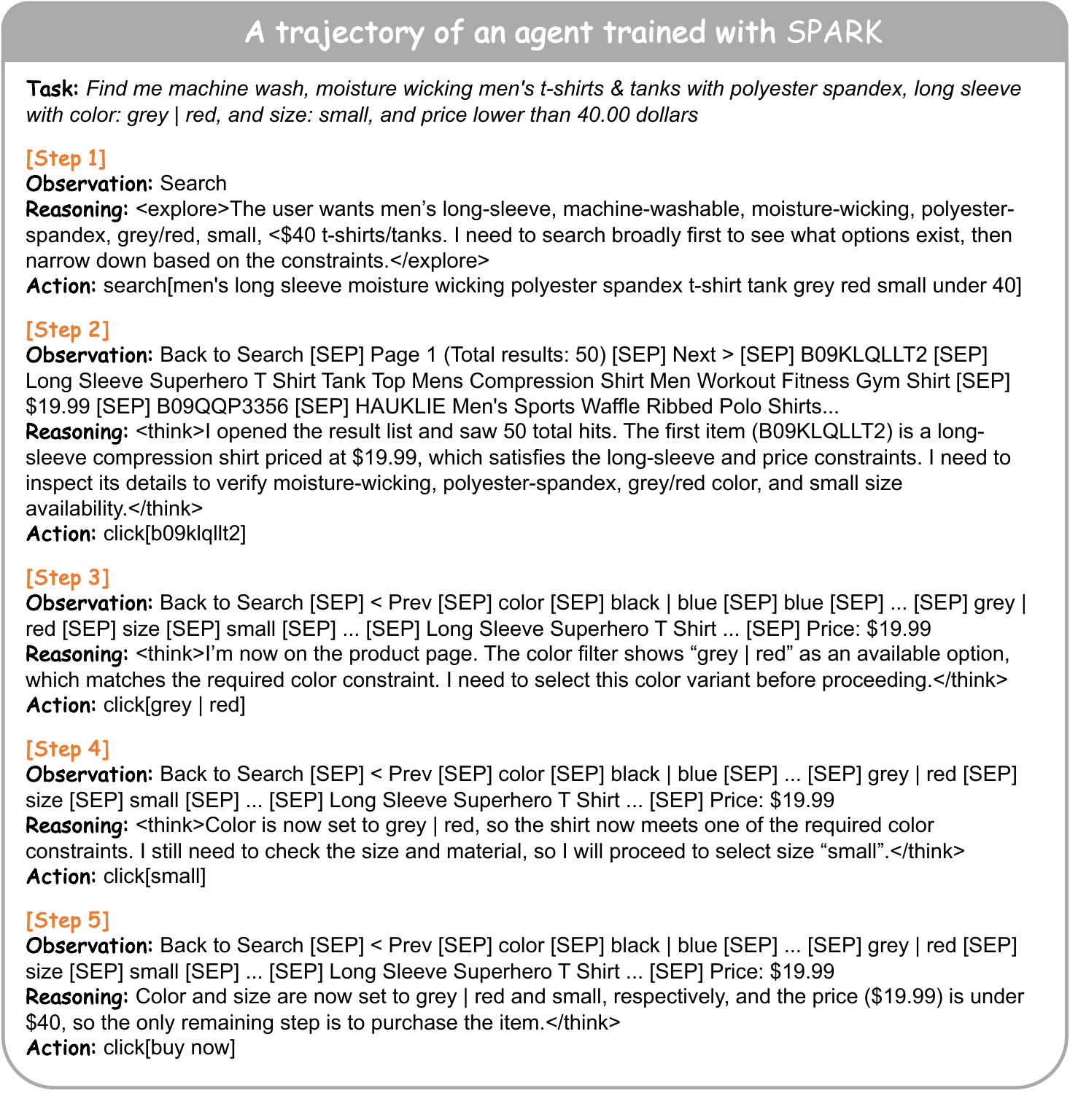}
    \caption{A full trajectory of \textsc{Spark} in WebShop (Example 4).}
    \label{fig:fullcase_webshop2}
\end{figure*}

Figures \ref{fig:fullcase_alfworld}–\ref{fig:fullcase_webshop2} provide illustrative examples from the ALFWorld and WebShop benchmarks.

\section{Additional Discussion}\label{subsec:discussion}
\subsection{Discussion on Tree-Based Baselines}
Tree-structured exploration methods have gained significant attention in recent reinforcement learning research~\citep{li2025treepo,wu2025templaterl,yang2025treerpo}. However, the majority of existing tree-based approaches, such as TreeRL~\citep{treerl}, have been primarily developed and evaluated within the math domain, where problem structures are relatively self-contained and solution verification is straightforward.

The long-horizon agentic tasks considered in our work present fundamentally different challenges~\citep{zhang2025landscape,singh2025agentic}. In embodied planning~\citep{shridhar2020alfworld} and web navigation~\citep{yao2022webshop} scenarios, trajectories often span 20-30+ sequential interactions with dynamic environments. Applying conventional tree-structured exploration to such settings would incur prohibitive computational overhead, as the branching factor compounds across the extended horizon. For instance, even a modest branching factor of 2 at each step would yield $2^{30} > 10^9$ potential trajectories for a 30-step episode, rendering exhaustive tree search practically infeasible. Furthermore, while some prior methods incorporate forms of adaptive branching, they typically rely on external signals (e.g., process reward models) or predefined heuristics that do not transfer well to open-ended agentic environments where dense supervision is unavailable. In contrast, \textsc{Spark} leverages intrinsic decision-making signals to selectively trigger branching only at critical states, achieving a principled balance between exploration depth and computational efficiency.

Given these considerations, we focus our empirical comparisons on chain-like RL methods (e.g., GRPO) that represent the current practical paradigm for long-horizon agent training. Adapting tree methods to long-horizon agentic domains remains an important direction for future work.

\subsection{Discussion on Related works}
Several concurrent works also explore uncertainty-aware or tree-structured rollout allocation. ARPO~\cite{dong2026agentic} targets distribution shifts after tool calls, while FR3E~\cite{zheng2025first} identifies high-entropy tokens to provide fine-grained feedback signals. DeepSearch~\cite{wu2026deepsearch} employs MCTS-based state-space coverage to overcome RL training plateaus. While these methods employ uncertainty-aware exploration, they fundamentally differ from \textsc{Spark} in both motivation and mechanism. First, in terms of motivation, ARPO and FR3E adopt a \emph{reactive, local} perspective---correcting errors after they occur---whereas \textsc{Spark} pursues \emph{proactive, strategic} resource allocation across the entire trajectory horizon. Second, in terms of mechanism, ARPO and FR3E rely on \emph{external statistical signals} (e.g., token-level entropy), treating the agent as a passive subject of statistical fluctuations; DeepSearch applies algorithmic heuristics (UCT scores) from an external controller. In contrast, \textsc{Spark} introduces \emph{intrinsic semantic reasoning} via the \texttt{<explore>} tag, enabling the agent to autonomously identify and act on its own epistemic uncertainty within its cognitive flow---transitioning from a passive subject of external measurement to an active, self-aware participant in exploration strategy. This intrinsic mechanism generalizes robustly to unseen domains, as evidenced by \textsc{Spark}'s strong out-of-domain performance (Table~\ref{tab:results}).

\subsection{Discussion on Reducing Dependence on Human Priors}
A key design principle of \textsc{Spark} is to minimize reliance on human-engineered heuristics for determining when and where to explore. Traditional exploration strategies often depend on handcrafted rules, domain-specific knowledge, or external reward models that require substantial human effort to design and may not generalize across tasks. In contrast, \textsc{Spark} enables the agent to autonomously identify critical decision points through its own reasoning process. The \texttt{<explore>} signal emerges from the agent's internal deliberation when it recognizes epistemic uncertainty or semantic ambiguity in the current state. This intrinsic mechanism offers two advantages: (1) it eliminates the need for task-specific annotations, and (2) it allows the exploration strategy to adapt naturally to novel scenarios, as the agent learns to recognize uncertainty patterns rather than relying on fixed rules.

Our experimental results support this claim. \textsc{Spark} demonstrates robust generalization on out-of-domain tasks (L2 splits), where human-defined heuristics would likely fail due to the absence of prior knowledge about unseen task categories. The agent's ability to autonomously trigger exploration at appropriate junctures without explicit supervision, validates that strategic exploration can emerge from learned intrinsic signals rather than external human guidance.

\subsection{Failure Mode Analysis}
While \textsc{Spark} demonstrates strong performance across diverse benchmarks, we identify several scenarios where it may underperform. First, when the base model has limited capability to recognize uncertainty, the \texttt{<explore>} signal may not be reliably triggered at genuinely critical states, leading to either missed exploration opportunities or spurious branching at routine steps. Second, in tasks where critical decisions are densely distributed throughout the trajectory, the advantage of selective branching diminishes, and \textsc{Spark} may not significantly outperform uniform exploration strategies. Future work could address these limitations by incorporating learned uncertainty estimators to calibrate branching decisions, or developing curriculum strategies that progressively refine exploration signals.

% \subsection{Future Works}

\end{document}